\documentclass[10pt,twocolumn,letterpaper]{article}

\usepackage[pagenumbers]{cvpr} 
\usepackage[dvipsnames]{xcolor}

\newcommand{\BO}[1]{{\boldsymbol{\rm #1}}}
\newcommand{\OP}[1]{{\operatorname{#1}}}

\definecolor{cvprblue}{rgb}{0.21,0.49,0.74}
\usepackage[pagebackref,breaklinks,colorlinks,citecolor=cvprblue]{hyperref}

\usepackage{cuted}
\usepackage{multirow}

\title{SparseDC: Depth Completion from sparse and non-uniform inputs}

\author{
Chen Long$^1$\thanks{These authors contribute equally to this work.}\quad
Wenxiao Zhang$^2$\footnotemark[1]\quad
Zhe Chen$^1$\quad
Haiping Wang$^1$\quad \\
Yuan Liu$^3$\quad
Zhen Cao$^1$\quad
Zhen Dong$^1$\thanks{Corresponding authors.}\quad
Bisheng Yang$^1$\footnotemark[2]\\
$^1$Wuhan University
$^2$Singapore University of Technology and Design
$^3$The University of Hong Kong\\
{\tt\small \{chenlong107,ChenZhe\_WHU,hpwang,zhen.cao,dongzhenwhu,bshyang\}@whu.edu.cn}\\
{\tt\small wenxxiao.zhang@gmail.com,yuanly@connect.hku.hk}
}

\begin{document}
\maketitle

\begin{strip}
    \centering
    \vspace{-5em}
    \centering
    \includegraphics[width=\linewidth]{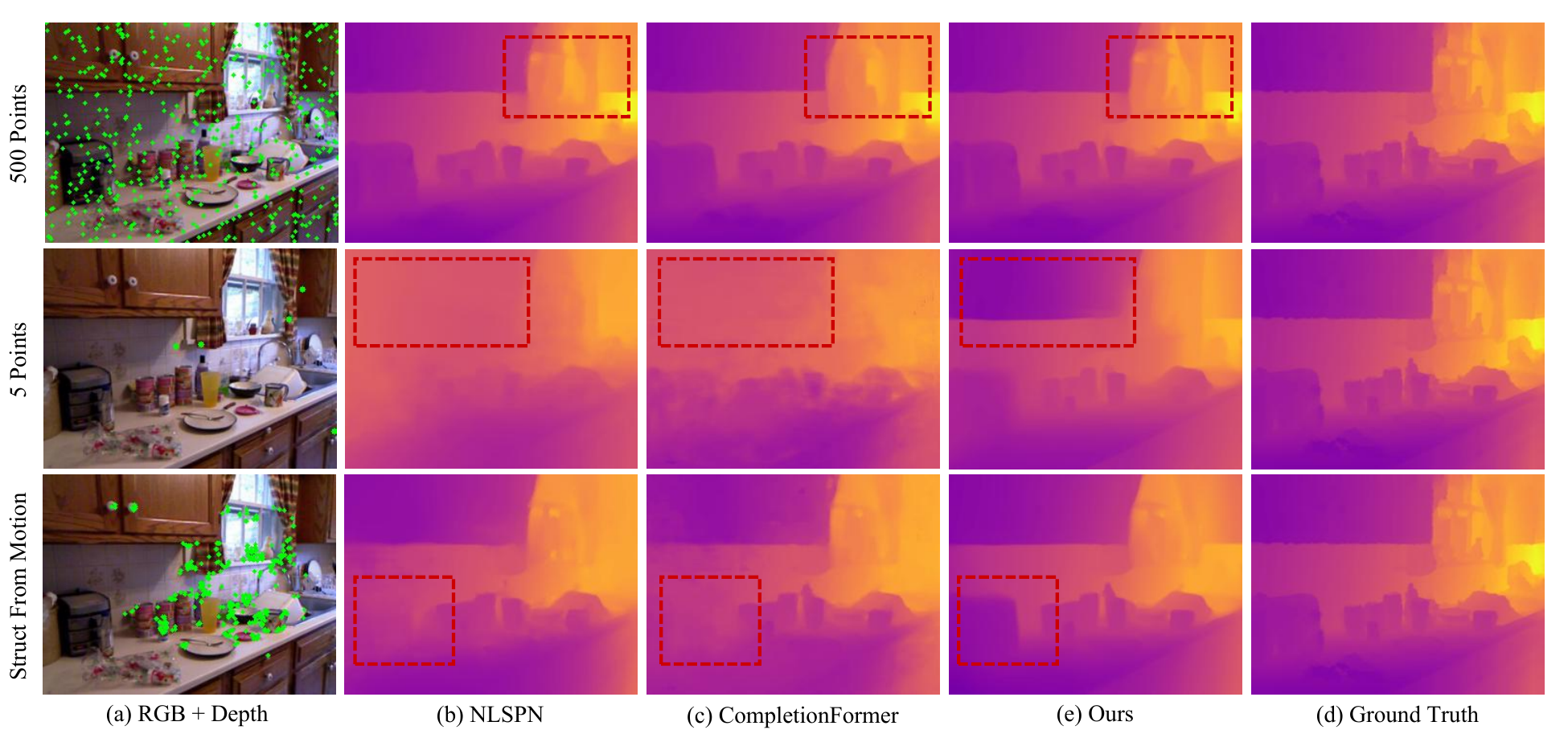}
    \vspace{-2em}
    \captionof{figure}{\textbf{Depth Completion from sparse and non-uniform inputs}. From top to bottom, we show the completed depth of state-of-the-art methods and ours using 500 depth points, 5 depth points, and non-uniform depth points obtained by SFM, respectively. Unlike the degradation of comparison methods, our framework can achieve reliable prediction under the above settings.}
    \label{fig:teaser}
\end{strip}

\begin{abstract}
We propose SparseDC, a model for \textbf{D}epth \textbf{C}ompletion of \textbf{Sparse} and non-uniform depth inputs. Unlike previous methods focusing on completing fixed distributions on benchmark datasets (e.g., NYU with 500 points, KITTI with 64 lines), SparseDC is specifically designed to handle depth maps with poor quality in real usage. 
The key contributions of SparseDC are two-fold.
First, we design a simple strategy, called SFFM, to improve the robustness under sparse input by explicitly filling the unstable depth features with stable image features.
Second, we propose a two-branch feature embedder to predict both the precise local geometry of regions with available depth values and accurate structures in regions with no depth. The key of the embedder is an uncertainty-based fusion module called UFFM to balance the local and long-term information extracted by CNNs and ViTs. Extensive indoor and outdoor experiments demonstrate the robustness of our framework when facing sparse and non-uniform input depths. 
The pre-trained model and code are available at \url{https://github.com/WHU-USI3DV/SparseDC}.

\end{abstract}
\section{Introduction}
\label{sec:intro}
Acquiring dense and accurate depth is crucial in various fields, such as autonomous driving\cite{yurtsever2020survey,maurer2016autonomous}, 3D reconstruction\cite{geiger2011stereoscan,long2022pc2,pan2016dense}, scene understanding\cite{cordts2016cityscapes,li2009towards}, AR/VR\cite{yuen2011augmented,park2019literature}. Although the commonly used depth sensors, such as Microsoft Kinect\cite{zhang2012microsoft}, Intel RealSense\cite{keselman2017intel}, LIDAR\cite{atanacio2011lidar}, can obtain depth within their effective field of view, it is still a great challenge to obtain dense depth maps due to environment, equipment, and cost limitations. In recent years, depth completion, serving as a cost-effective means to achieve dense pixel-wise depth, has become a long-standing research endeavor that has captivated the attention of many researchers.

Despite achieving impressive performance on benchmark datasets\cite{nyu,kitti}, existing state-of-the-art methods\cite{wang2023lrru,youmin_completionformer_2023,yan_rignet_2022,zhou_bev_2023} often fall short in some challenging cases.
In contrast to the fixed data distribution of the benchmark dataset(e.g., NYU Depth\cite{nyu} with 500 points, Kitti\cite{kitti} with 64 lines), the inputs in real scenarios may be non-uniform and sparser. For example, researchers often employ depth sensors with lower resolutions (Velodyne HDL-16E LiDAR, VCSEL ToF sensors) to mitigate costs. In addition, environmental factors easily influence sensors, such as high-reflectivity objects and multipath effects resulting in uneven distribution of effective depth points. Regrettably, mainstream methods primarily focus on the sparse-to-dense completion process without realizing the sparsity changes and the non-uniform distribution of depth points. Consequently, they are difficult to reconstruct the scene structure when the spatial pattern of the input significantly deviates from the distribution observed in the benchmark dataset during training, as illustrated in \cref{fig:teaser}

The sparse and non-uniform input depth brings great challenges in extracting stable and robust features for the completion task. In the regions with dense input depth, the network needs to learn to trust the local dense input depth values while in regions with sparse or even no depth, the network is supposed to speculate the depth values from the RGB features and other long-term context information. However, the changing input depth patterns shown in \cref{fig:teaser}(a) make the network confusing in which kind of features should be extracted even with data augmentation to simulate the changing depth in training~\cite{woo2023convnext,graham2017submanifold,spconv2022,Sparsitycnn}. To solve this problem, we introduce \textit{Sparse Feature Filling Module} that explicitly fills the unstable depth features with stable image features thereby improving the robustness of the network to various spatial distributions of input depth.

\begin{figure}[t]
    \centering
    \includegraphics[width=1.0\linewidth]{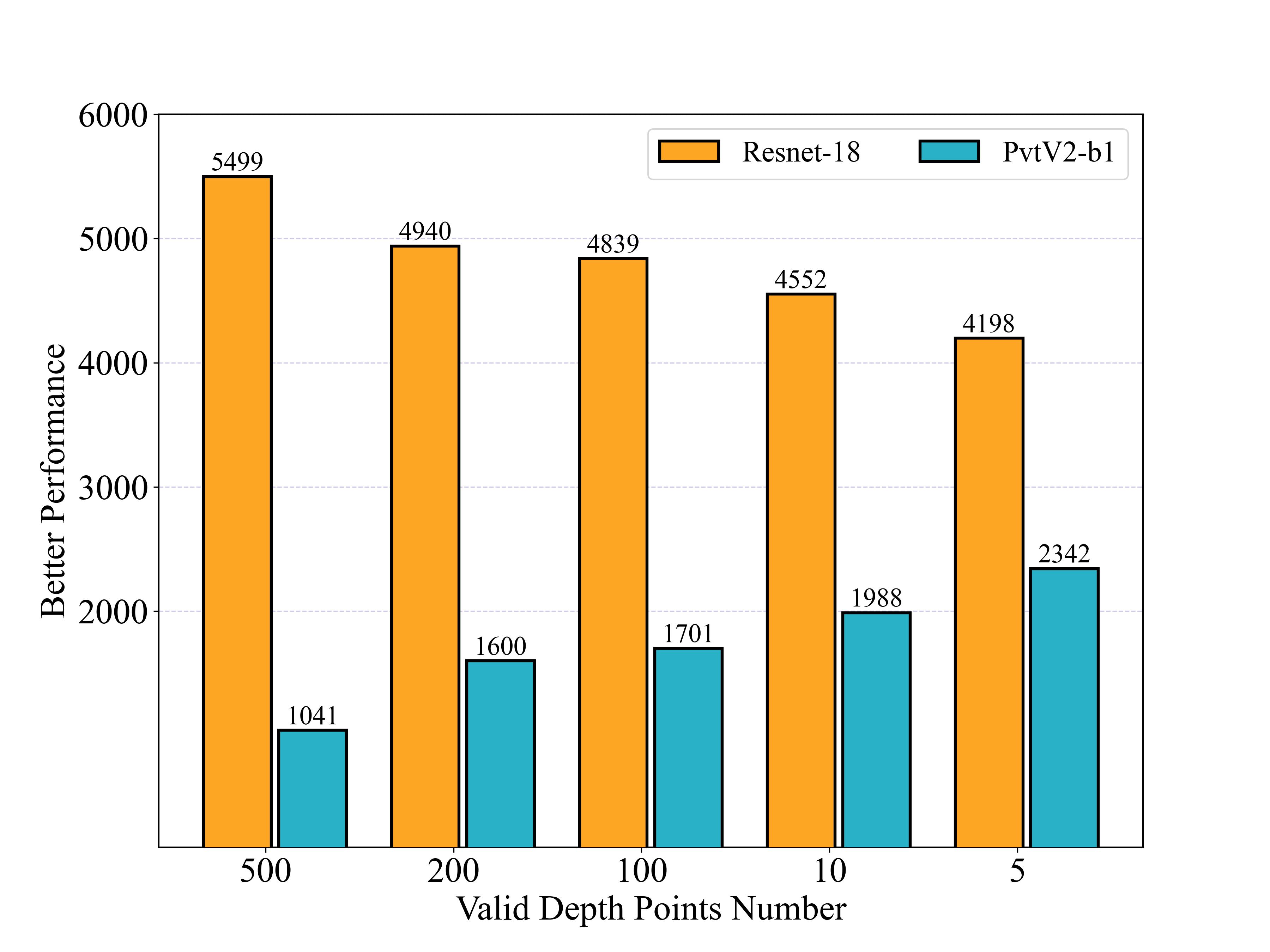}
    \caption{\textbf{Pilot study results on NYU Depth V2 dataset}. The horizontal axis represents the number of effective depth points in each testing depth map. The vertical axis illustrates the number of testing depth maps for which the network outperforms the other in terms of RMSE.}
    \label{anaylse}
    \vspace{-1em}
\end{figure}

Furthermore, we observed that CNNs and ViTs show different characteristics in dealing with non-uniform input depths.
In a simple experiment, we select ResNet-18\cite{resnet} and PvtV2-b1\cite{pvt, pvtv2} as two backbone networks for the depth completion task and evaluate them on the NYU Depth V2\cite{nyu} with input depth maps with varying densities. As illustrated in \cref{anaylse}, though the overall performance of ResNet is better, as the density of input depth decreases, more and more depth values predicted by the PvtV2-b1 are more accurate than those predicted by the ResNet-18.
The reason is that high-density depth points are sufficient to provide rich local context and detail for depth completion, while for low-density inputs, we need to understand the overall scene to fill large areas. Correspondingly, CNNs are known for their ability to extract local features with fine-grained details while ViTs focus on capturing the global scene context to leverage global relationships among pixels. The long-term context helps the ViT to account for the sparse input depth values. This observation motivates us to design a two-branch network structure that consists of both CNNs and ViTs to combine the advantages of both parts for a better depth completion on the sparse and non-uniform input depth maps.

An important problem here is how to fuse the two-branch features to get robust features. As discussed before, the network needs to put different importance on features from the CNNs and ViTs according to the local density of the input depth maps. However, a naive fusion~\cite{youmin_completionformer_2023,wang2023lrru}, which simply concatenate or add the features from two branches, actually treats features from two branches equally and thus leads to degenerated performances.
To address this problem, we design an \textit{Uncertainty-Guided Feature Fusion Module(UFFM)} to dynamically fuse extracted features explicitly. The UFFM uses sparse depths to estimate and correct the uncertainty of each pixel and utilizes this information to guide the fusion process, enabling the network to adjust the relative contribution of each branch based on the distribution patterns of the inputs.

\begin{figure*}[t]
    \centering
    \includegraphics[width=0.8\linewidth]{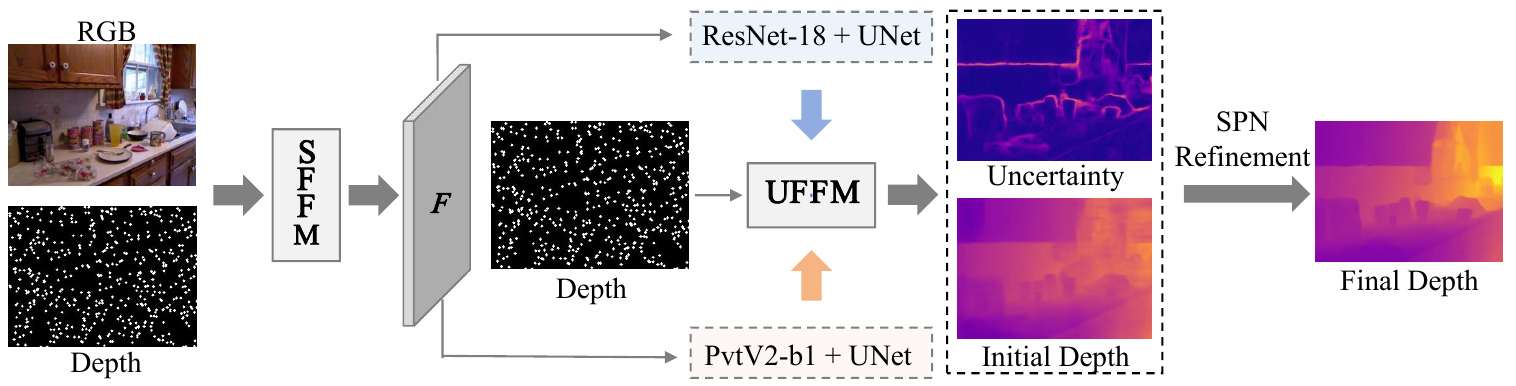}
    \caption{\textbf{SparseDC Architecture.} Given a sparse depth map and an image, we obtain a stable initial feature $\BO F$ using SFFM. Then, we adopt a two-branch feature extractor to get multi-scale features. After that, the extracted features and sparse depth map are fed to UFFM to obtain the initial depth and estimated uncertainty. Finally, a non-local propagation network\cite{vedaldi_non-local_2020} is exploited for final refinement.}
    \label{fig:overview}
    \vspace{-1em}
\end{figure*}

In summary, we propose a novel deep completion framework named SparseDC, \cref{fig:overview} illustrates the whole pipeline with above new components. SparseDC exhibits significant robustness improvement with better efficiency when dealing with sparse and non-uniform depths. We conduct a comprehensive experimental evaluation of our approach on NYU Depth\cite{nyu}, KITTI DC\cite{kitti}, and SUN RGB-D\cite{song2015sun} and demonstrate 17\% improvements on REL metric, 7.8\% improvements on RMSE metric, respectively, over the State-of-the-art method CompletionFormer\cite{youmin_completionformer_2023}. 
Overall, our contributions can be summarized as follows:

\begin{itemize}
    \item We analyze the limitations of mainstream methods when dealing with sparse and non-uniform inputs and propose a novel depth completion framework called SparseDC to tackle this. It consists of several modules which can help extract stable and robust features for the completion task.
    \item We access the performance between ours and state-of-the-art methods on some simulated and real challenging cases using indoor\cite{nyu,song2015sun} and outdoor datasets\cite{kitti}. Extensive experiments show the robustness of SparseDC when dealing with sparse and non-uniform depths. 
\end{itemize}

\section{Related Work}
\label{sec:rw}
\textbf{Depth Completion} is a fundamental task in computer vision. It aims at recovering dense pixel-level depth from sparse depth maps obtained from sensors. In recent years, deep learning-based methods have emerged as the leading approaches to this task. 
Uhrig et al.\cite{Sparsitycnn} introduced sparse convolution as an alternative to traditional convolution and established benchmark datasets based on Kitti for evaluating depth completion algorithms. Later, Ma et al.\cite{sparse2dense} incorporated RGB data as an additional input, significantly improving the performance for the deep completion task. Building upon this\cite{sparse2dense}, researchers have been growing interest in exploring multimodal data fusion in their frameworks, such as images\cite{metzger_guided_2022,rho_guideformer_2020,jeon_abcd_2022,liu_learning_2021,zhao_adaptive_2021,deeplidar}, normals\cite{xu2019depth,gu_denselidar_2021}, semantics\cite{markert_segmentation-guided_2022}, and residual depth maps\cite{liu_fcfr-net_2020}. Then, Cheng et al. proposed a convolutional spatial propagation network CSPN\cite{ferrari_depth_2018}, the first to apply spatial propagation techniques to depth completion. After that, methods such as CSPN++\cite{cheng_cspn_2019}, NLSPN\cite{vedaldi_non-local_2020}, DySPN\cite{lin_dynamic_2022}, GraphCSPN\cite{liu_graphcspn_2022}, etc. have been proposed, they improve the accuracy and efficiency of depth completion by designing different convolution strategies (deformable convolution, graph convolution, etc.) to propagate the local information provided by the depth points over image, and achieve SOTA performance on benchmark datasets. To push the envelope of depth completion, recent approaches tended to use complex network structures and sophisticated learning strategies\cite{hu_deep_2023,zhou_bev_2023,wang2023lrru,yan_rignet_2022}, such as BEV@DC\cite{youmin_completionformer_2023}, which proposed a more efficient and powerful multi-modal training scheme to boost the performance of image-guided depth completion. However, these methods mainly emphasize sparse-to-dense conversion without realizing the impact of sparsity changes and non-uniform depth points, which limits their scope of applications in real scenarios.

Some researchers have considered this problem and given their solutions. Andrea Conti et al. proposed SpAgNet\cite{SpAgnet}. The key idea is not to directly feed sparse depth points to the convolutional layers, but to iteratively merge the sparse input points with multiple depth maps predicted by the network during the decoding process. This approach mitigates the effects of uneven inputs. However, the receptive field of the depth points is very restricted, leading to the final result that relies on the performance of the backbone network, which increases the number of parameters of the model (51M $\gg$ 25.8M NLSPN\cite{vedaldi_non-local_2020}) and leads to a domain bias, limiting its applicability. Another related research is Sparse SPN\cite{sparsespn}, which propagates sparsely distributed depth values to the rest of the image by designing a cross-shaped spatial propagation network. However, this approach still relies on the local inductive bias provided by the spatial propagation technique, and has limited ability to model depth values over long ranges. In addition, Sparse SPN\cite{sparsespn} is designed only for keypoint sampling, which cannot handle other spatial patterns such as uneven density and big holes.

\section{Method}
\label{sec:method}
\subsection{Overview}
\label{sec:3.1}
Given a sparse depth map $\mathcal{S} \in \mathbb{R}^{H\times W \times 1}$ and an image $\mathcal{I} \in \mathbb{R}^{H\times W \times 3}$, our goal is to restore a dense pixel-wise depth $\mathcal{D} \in \mathbb{R}^{H\times W \times 1}$. Different from mainstream methods, we focus on dealing with \textbf{\textit{sparse}} and \textbf{\textit{non-uniform}} inputs.

The whole pipeline of SparseDC is illustrated in \cref{fig:overview}. First, we feed the depth map $\mathcal{S}$ and image $\mathcal{I}$ into the \textit{SFFM} to obtain an initial feature map $\BO{F} \in \mathbb{R}^{H\times W \times C}$. Then, we embed $\BO{F}$ with two separate feature extractors, composed by CNNs and ViTs, to respectively extract so-called local multi-scale features $\BO{F}_{local}=\{{f}^n_{l}\}_{n=0}^{N}$ and global multi-scale features $\BO{F}_{global}=\{{f}^n_{g}\}_{n=0}^{N}$, where $f_*^n$ indicates the $n^{th}$ scale feature map.
After that, we design an \textit{UFFM} to predict the pixel-wise uncertainty to guide the fusion of local features $\BO{F}_{local}$ and global features $\BO{F}_{global}$ at different scales. We thus get the fused features $\BO{F}_{fuse} =\{{f}^n\}_{n=0}^{N}$ to estimate the initial depth map and uncertainty. Finally, we apply the non-local spatial propagation module\cite{vedaldi_non-local_2020} to refine the initial depth map and get the final depth $D$.

\subsection{Sparse Feature Fill Module}
\label{sec:3.1}
\begin{figure}[!h]
    \centering
    \includegraphics[width=0.8\linewidth]{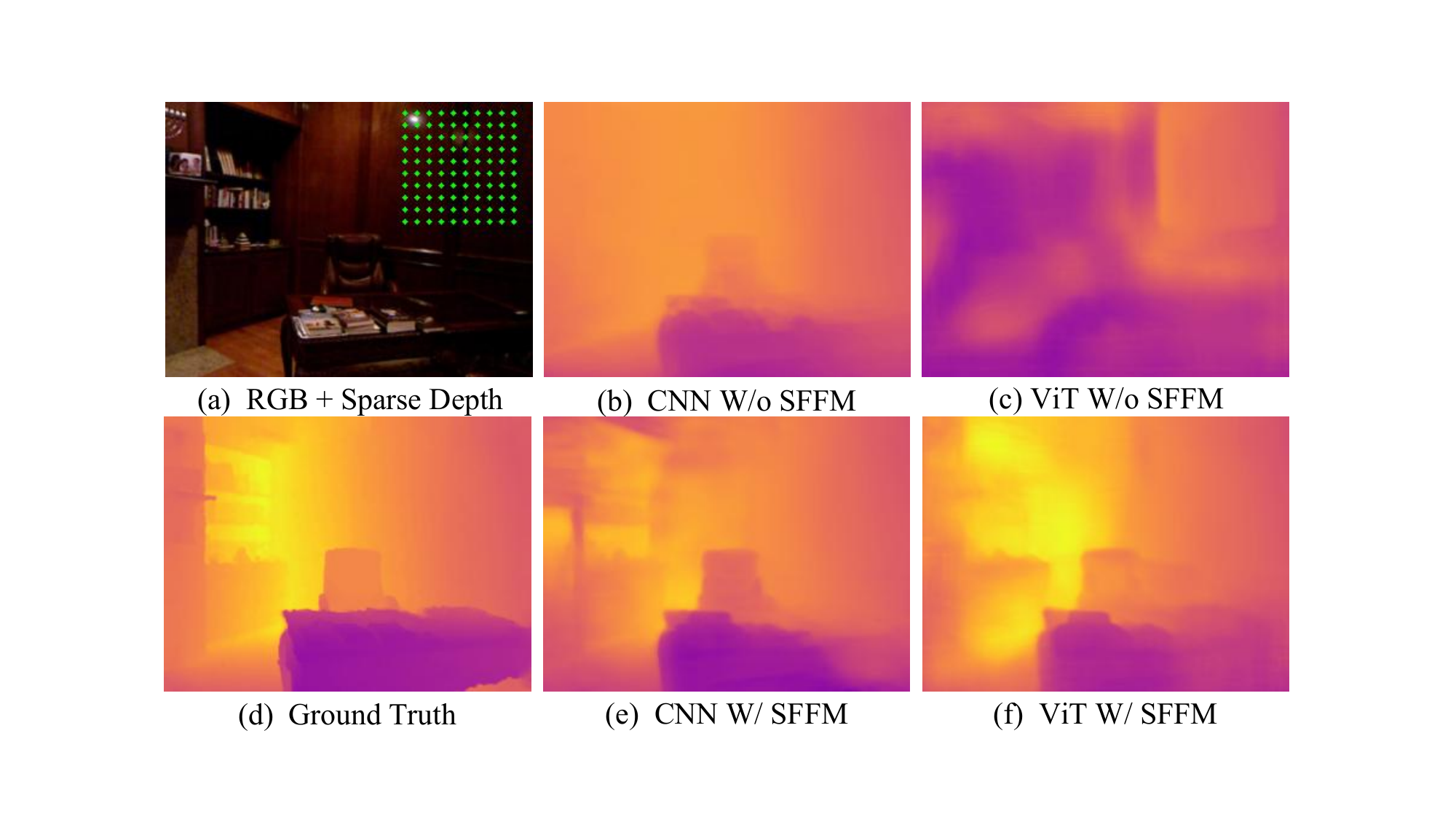}
    \caption{\textbf{Visual comparison.} Illustration of the effectiveness of our proposed SFFM on CNN and ViT backbones with non-uniform depth inputs.}
    \vspace{-1em}
    \label{fig:sffm_comparsion}
\end{figure}

Given an image and a depth map, SFFM aims to output a feature map mitigating the feature instability\cite {woo2023convnext} caused by non-uniform depths.
As shown in \cref{fig:sffm_comparsion}, existing methods struggle to maintain the geometry shape in the face of non-uniform inputs, which can be alleviated by SFFM.

\begin{figure}[t]
    \centering
    \includegraphics[width=1.0\linewidth]{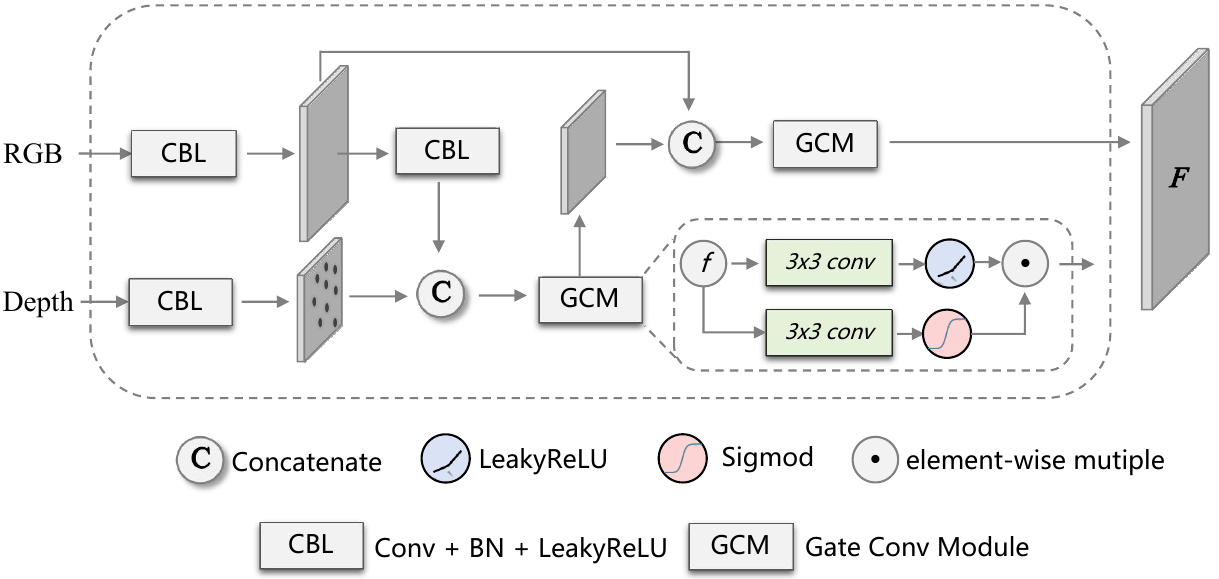}
    \caption{\textbf{An illustration of SFFM.} SFFM aims to output stable and robustness features for depth completion.}
    \label{fig:sffm}
    \vspace{-1em}
\end{figure}

\cref{fig:sffm} depicts the detailed structure of the SFFM. Given the sparse and non-uniform depth map $\mathcal{S}$ and the image $\mathcal{I}$, we first extract the features of the image and depth with a stack of 2D convolution, batch normalization\cite{batchnorm}, and LeakyReLU\cite{relu,leakyrelu} layer. Then, we concatenate the image and depth features on the feature dimension. 
After that, we use gate convolution module\cite{gate} with channel attention to fuse these features. It learns the weights of the stable image features and sparse depth features based on the distribution of input depths, and use the weight to dynamically guide the feature filling to obtain $\BO F_{dep}^{'}$. Finally, $\BO F_{dep}^{'}$ and $\BO F_{rgb}$ are concatenated to obtain a stable feature input $F$ by using another gated convolution layer.
In addition, to control the feature range of $F$ and accelerate the convergence of the model, we predict a coarse depth map $D^{'}$ by feeding $F$ to an extra depth decoder head for intermediate supervision.


\subsection{Two Branch Feature Extraction}
Motivated by the observations in \cref{anaylse}, we introduced a two-branch feature extraction architecture, one convolution-based branch focuses on extracting the local information provided by the depth points, called Local Branch, and the other branch is based on the vision transformer to provide indispensable global information, called Global Branch. 

\textbf{Local Branch}. We choose ResNet-18\cite{resnet} pretrained on ImageNet\cite{imagenet} as the backbone network of the convolution branch. Specifically, it downsamples the feature map $\BO F$ to scales $\frac{1}{1},\frac{1}{2},\frac{1}{4},\frac{1}{8},\frac{1}{16}$. After that, the features will be used in the UNet\cite{unet} decoding step as input to get $\BO F_{local}$. 

\textbf{Global Branch}. We chose the lightweight PvtV2-b1\cite{pvtv2} also pre-trained on ImageNet\cite{imagenet} as our global feature extract backbone. This branch first splits the input Feature $\BO F$ into $4\times4$ patches by a convolution filter with a kernel size of 7. After that, a linear embedding layer is applied to project the downsampled feature to an arbitrary dimension. Then, it uses 4 transformer layers to downsample the initial feature to $\frac{1}{1},\frac{1}{4},\frac{1}{8},\frac{1}{16},\frac{1}{32}$, and these features will also be used in decoding steps to get $\BO F_{global}$, like $\BO{F}_{local}$. 

\subsection{Uncertainty-Guide Feature Fusion Module}
Given non-uniform and sparse inputs, the network needs to learn to trust the local dense input depth values in the regions with dense input depth. In regions with sparse or even no depth, the network is supposed to speculate the depth values from long-term context information. 
\begin{figure}[!h]
    \centering
    \includegraphics[width=1.0\linewidth]{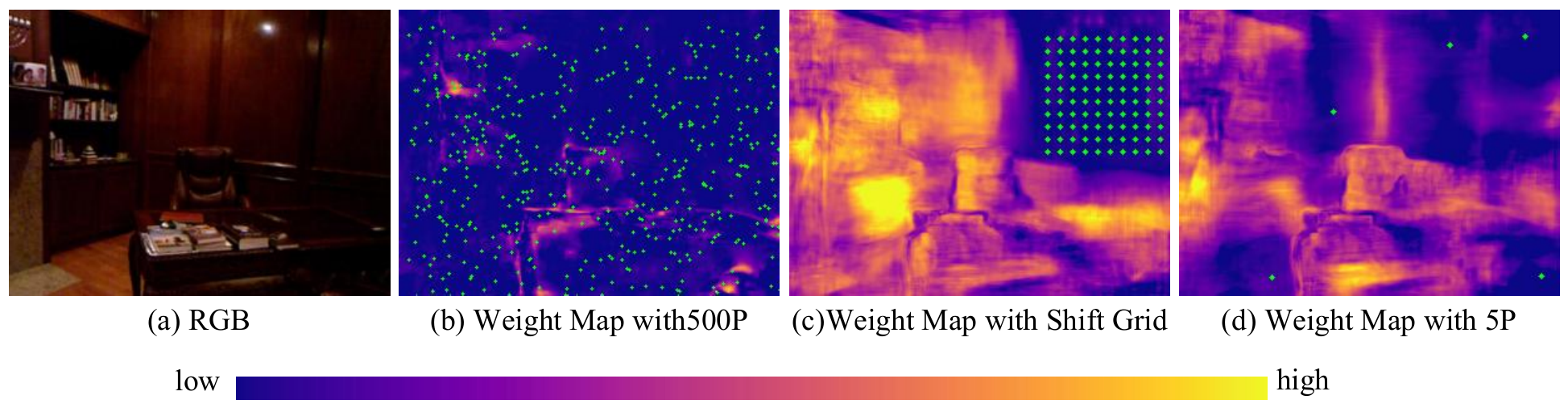}
    \caption{\textbf{Weight Map Visualization.} We visualize the weight map with different input patterns. The darker the color, the more the global branch contributes to the whole model.}
    \label{fig:weight}
    \vspace{-1em}
\end{figure}

To achieve this, we propose a weighted fusion strategy that enables the network to dynamically determine the confidence levels of local and global branch features.
Specifically, we predict the pixel-wise uncertainty of each feature branch. Then, we use the input depth points to rectify the uncertainty and finally use uncertainty to guide the explicit fusion of features. \cref{fig:weight} illustrates the effect of UFFM in the face of sparse and non-uniform inputs. UFFM can adjust the contributions of each branch according to the inputs' spatial pattern, realizing the complementary advantages of global and local information.
 \begin{figure}[!h]
    \centering
    \includegraphics[width=1.0\linewidth]{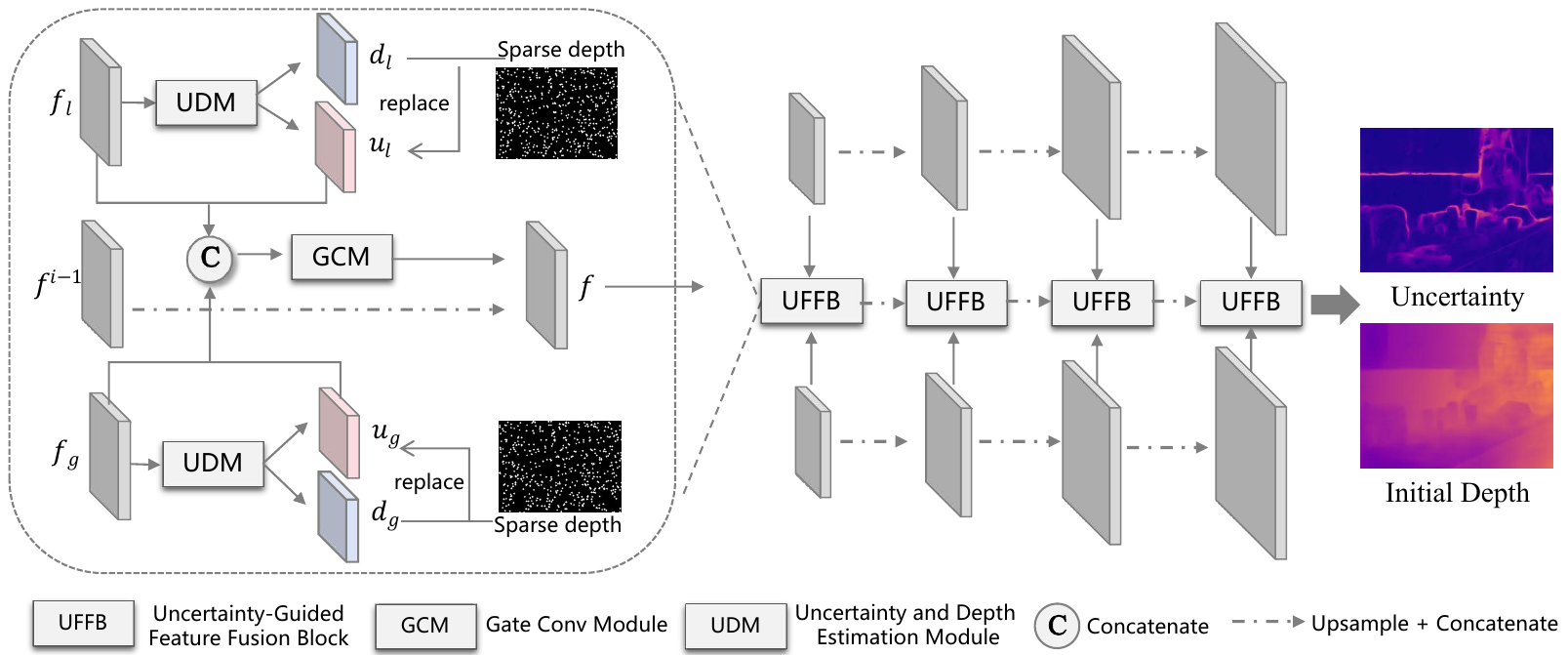}
    \caption{\textbf{UFFM Architecture.} UFFM aims to dynamically fuse the extracted features and then estimate the initial depth and reliable uncertainty.}
    \label{fig:uffm}
    \vspace{-1em}
\end{figure}

In practice, UFFM applies several Uncertainty-guided Feature Fusion Block (UFFB) at different scales $\frac{1}{1},\frac{1}{2},\frac{1}{4},\frac{1}{8}$. \cref{fig:uffm} illustrates the detailed structure of UFFM. Given the multi-scale local features $\BO F_{local}$ and global features $\BO F_{global}$, UFFB first uses two different convolutional layers (depicted as UDM in \cref{fig:uffm}) to compute an uncertainty and a depth map at this scale. These will be fed to the UFFB of the next scale to correct the predicted results iteratively. 
Since there is no ground truth for the uncertainty prediction, we compute a pseudo ground truth uncertainty inspired by \cite{shao2023urcdc}, formulated in \cref{eq:uncertainty}.
\begin{equation}\label{eq:uncertainty}
    \hat{u} = 1 - \OP {exp}(-\frac{|d^n - g^n|}{b \times g^n})
\end{equation}
where, $\hat{u}$ denotes the defined pseudo ground truth uncertainty, $g^n$ is the ground truth depth map at scale $n^{th}$, the $d^n$ is predicted depth maps. $b$ is a coefficient that controls the tolerance for error, we set $b$ as 0.1 in this paper to amplify differences between features. 
In addition, to further exploit the depth information, we downsample the sparse depth map $\mathcal{S}$ by sparsity aware pooling to get multi-scale depth maps. After that, we use \cref{eq:uncertainty} to calculate the accurate uncertainty and replace the estimated uncertainty at positions that have valid depth points. 
Then, we use the predicted and replaced uncertainty to guide the fusion of different features as shown in \cref{eq:fusion}.
\begin{equation}\label{eq:fusion}
    f^n = Gate((1 - u^n_{l}) \cdot f^n_{l} \oplus (1 - u^n_{g}) \cdot f^n_{g})
\end{equation}
where, $u^n_{l}$ denotes the local feature uncertainty, the $u^n_{g}$ denotes the global feature uncertainty, and the $f^n$ is the fused feature which will be fed to non-local spatial propagation module\cite{vedaldi_non-local_2020} to compute the final completed depth $D$. 
\subsection{Loss Fuction}
We train our network by supervising all the outputs of each UFFM. First, we downsample $G$ through average pooling to obtain multi-scale ground truths dense depth maps $\{g^n\}_{n=0}^{N}$ and use \cref{eq:uncertainty} to compute pseudo uncertainty. Then, we use $\mathcal{L}_1$ loss to supervise the predicted uncertainty and $\mathcal{L}_2$ loss for depth at each scale, which is described by \cref{eq:loss}.
\begin{equation}\label{eq:loss}
\begin{split}
    L_{g}^n &= 0.5|u^n_{g} - \hat{u}^n_{g}| + \|d^n_{g} -g^n\| \\
    L_{l}^n &= 0.5|u^n_{l} - \hat{u}^n_{l}| + \|d^n_{l} - g^n\| \\
    L^n &= 0.5|u^n - \hat{u}^n| + \|d^n - g^n\| \\
    L_{rec}^n &= 0.5(L_{g}^n + L_{l}^n) + L^n \\
\end{split}
\end{equation}
Then, we need to supervise the output of SFFM and non-local spatial propagation module, the overall loss function can be formulated by \cref{eq:overall}
\begin{equation}\label{eq:overall}
    L = \sum^N_{n=0} \gamma^n L_{rec}^n + 0.05\|D^{'} - G\| + |D - G| + \|D - G\|
\end{equation}
where $\gamma$ denotes the exponential decay factor, we set it as 0.8 in this paper to make the network pay more attention to the high-resolution outputs.
\section{Experiments}
\label{sec:Exp}
\subsection{Experimental Settings}
\label{sec:dataset}
In our experiments, we included both indoor and outdoor datasets.

\textbf{NYU Depth V2.} The NYU Depth V2\cite{nyu} is a widely used indoor dataset in depth completion, which contains 40,800 RGBD images captured by Microsoft Kinect with an original resolution of 640$\times$480. Each image and depth map should be down-sampled to 320$\times$240, then center-cropped to 304$\times$228. We follow the previous work\cite{sparse2dense} to divide the dataset and set the upper bound of depth to 10 meters. Unlike previous methods that randomly sampled 500 points from the ground truth as sparse input, we sampled 5-500 random points during training.

\textbf{SUN RGB-D} The SUN RGB-D dataset\cite{song2015sun} contains 10,335 raw RGB-D images captured by four different sensors. This dataset has more challenging scenes and sensors, which helps to evaluate model generalization efficiently.

\textbf{KITTI depth completion.} The KITTI dataset\cite{kitti} contains over 93,000 raw sparse LIDAR scans (retrieved by Velodyne HDL-64E LiDAR sensor) and corresponding RGB images. The training, validation, and test set have 86,000, 7,000, and 1,000 samples, respectively\cite{hu_deep_2023}, and the upper depth bound is 80 meters.
For outdoor acquisition scenarios, in addition to high-emissivity objects and noise, the other factor is the different vertical resolution of the different sensors, e.g., 64-line LIDAR vs. 16-line LIDAR. So, we devised a simple but useful training strategy on KITTI that randomly masks out all depths from 0\%-95\% of rows during training.

\textbf{Evaluation metrics.}
We follow \cite{hu2019revisiting,jiang2021plnet} to use three metrics for the dense depth prediction evaluation  \textbf{RMSE}: Root mean squared error, \textbf{MAE}: Mean absolute error,
\textbf{REL}: Mean relative error.
Performances using other metrics are reported in the Appendix.

\textbf{Baselines.} We compared SparseDC with serval state-of-the-art depth completion networks, including NLSPN (ECCV 2020)\cite{vedaldi_non-local_2020}, GraphCSPN (ECCV 2022)\cite{liu_graphcspn_2022}, CompletionFormer (CVPR 2023)\cite{youmin_completionformer_2023}, PENet (ICRA 2021)\cite{hu_penet_2021}. For a fair comparison, we retrained all of them using their code and our training strategy. Note that, we did not compare it with SpAgNet\cite{SpAgnet}, SparseSPN\cite{sparsespn} and some other state-of-the-art methods\cite{zhou_bev_2023,wang2023lrru,lin_dynamic_2022,yan_rignet_2022}, et al. because they did not release their code.

\begin{figure}[h]
    \centering
    \includegraphics[width=0.8\linewidth]{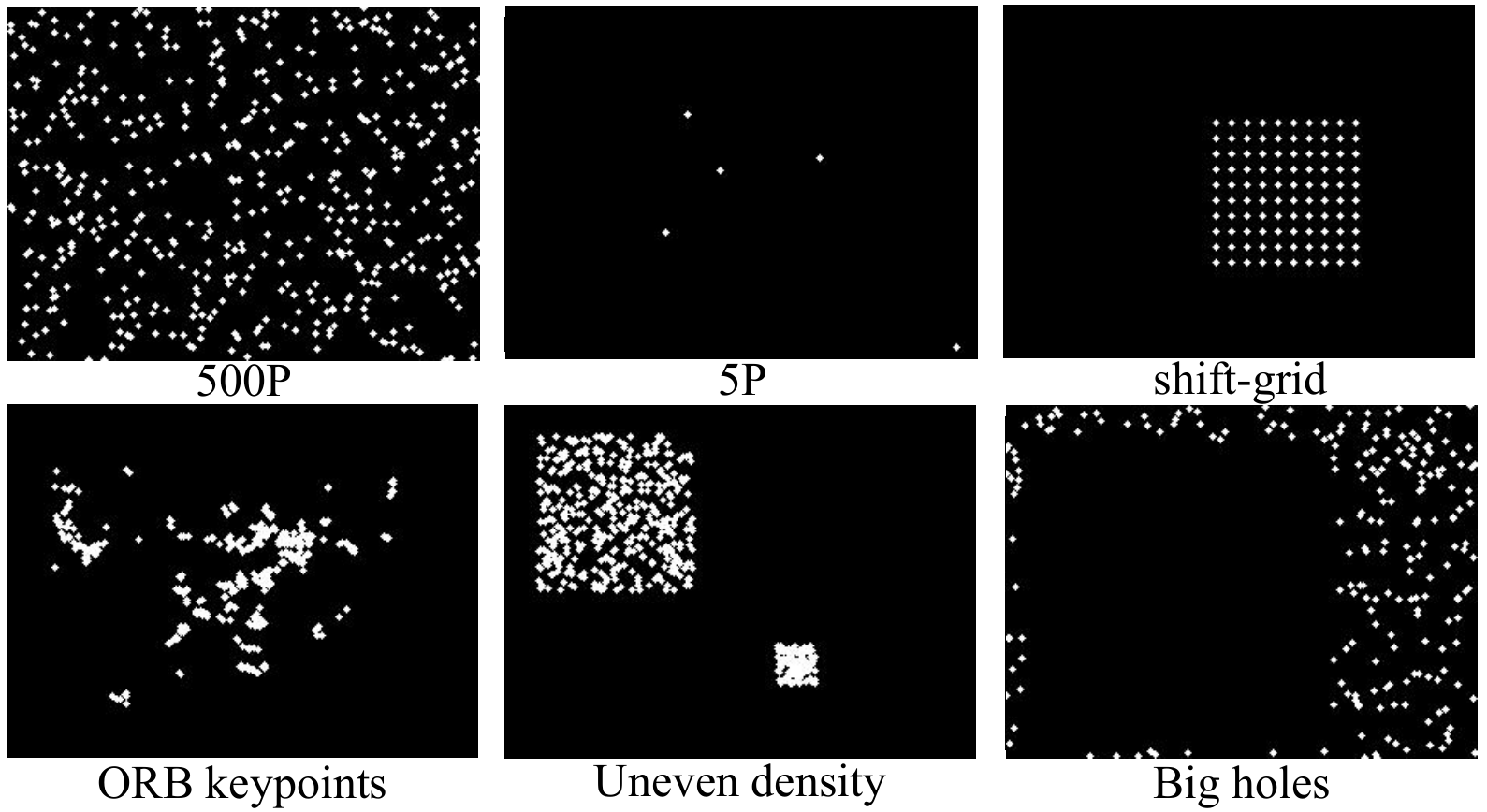}
    \caption{\textbf{Sparse depth patterns.} Including varying sparsity depth points, shift grid patterns (VCSEL TOF sensors), keypoints-based sampling, uneven density depth points, and big holes.}
    \vspace{-1em}
    \label{patterns}
\end{figure}

\textbf{Implementation details.} We trained our network with a batch size 16 for NYU (8 for each GPU) and 6 for KITTI (2 for each GPU) on the PyTorch platform, respectively. The Adam optimizer is used with an initial learning rate of $10^{-4}$, which is decreased in every unimproved five epochs by a decay rate of $0.3$. Our training process will stop when there is no improvement for ten consecutive epochs. We will release the source code and the trained models upon acceptance.

\begin{figure*}[t]
    \centering
    \includegraphics[width=1.0\linewidth]{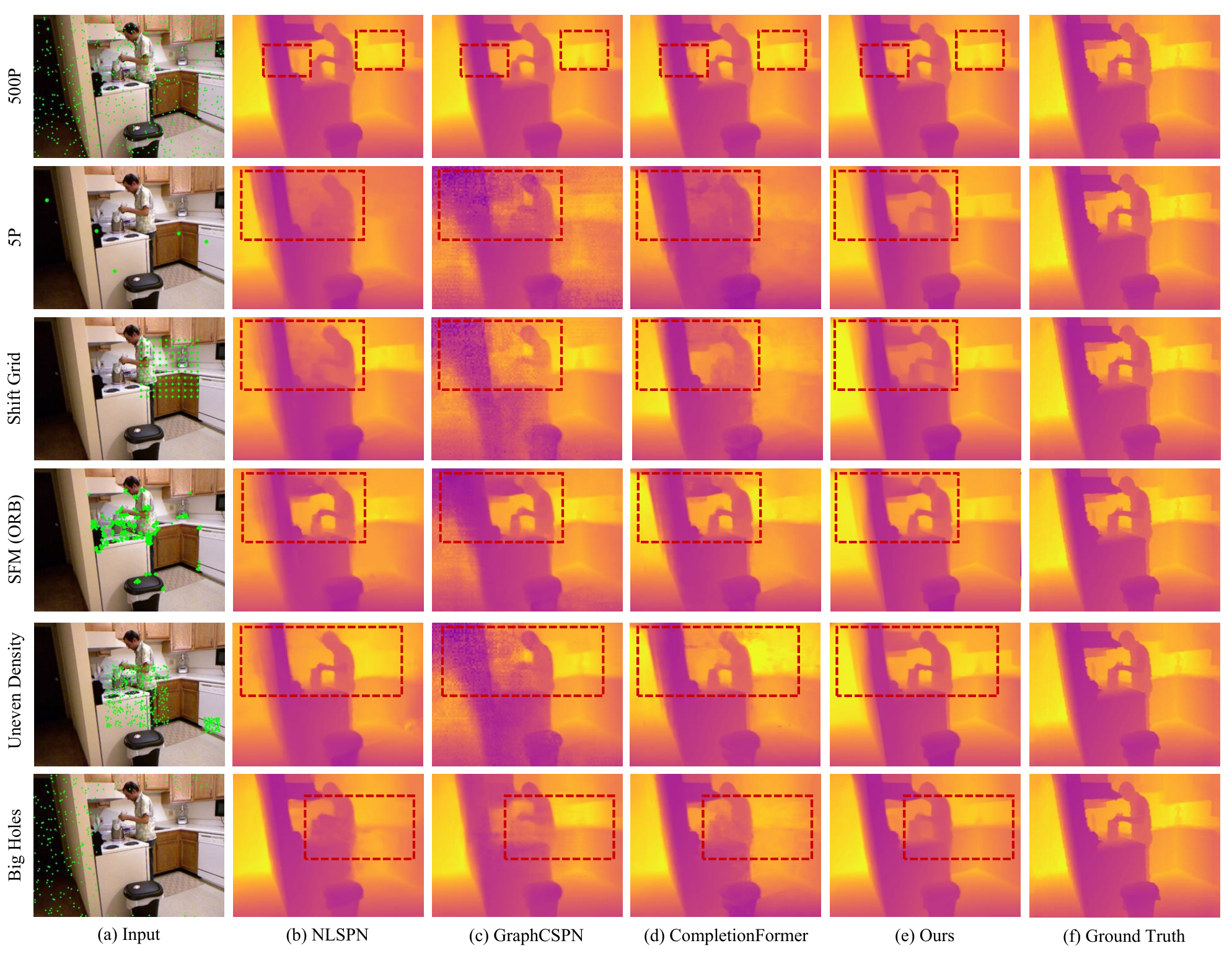}
    \caption{\textbf{Qualitative results on NYU Depth V2.} Comparing completion results with different patterns of inputs using different methods.}
    \label{fig:nyu}
    \vspace{-1em}
\end{figure*}

\subsection{Results on NYU Depth V2}
Unlike existing methods, which mainly focus on sparse-to-dense solutions, our networks pay more attention to dealing with sparse and non-uniform inputs. We trained our network on NYU Depth V2, and randomly sampled $5 \sim 500$ depth points as inputs during training. SpAgNet\cite{SpAgnet} has proved that existing methods trained on fixed spatial patterns(500 random samples) are not comparable in the face of some challenging cases. For example, with only 5 points, the RMSE by NLSPN\cite{vedaldi_non-local_2020} is \textbf{1.033m}. \textit{So, we retrained the comparison methods using their released code and the same training strategy with us for a fair comparison.}
\begin{table}[t]
    \centering
    \footnotesize
    \begin{tabular}{c|c|cc}
    \toprule
Model                                       & Samples                        & RMSE (m)$\downarrow$                                   & REL$\downarrow$                                    \\ \midrule
NLSPN\cite{vedaldi_non-local_2020}                           & 500                               & 0.0984                                 & 0.0136                                 \\
GraphCSPN\cite{liu_graphcspn_2022}                       & 500                               & 0.0990                                 & 0.0136                                 \\
CompletionFormer\cite{youmin_completionformer_2023}                & 500                               & 0.0984                                 & 0.0135                                 \\
\textbf{Ours}        & 500          & {\color[HTML]{FF0000} \textbf{0.0976}} & {\color[HTML]{FF0000} \textbf{0.0128}} \\ \midrule
NLSPN\cite{vedaldi_non-local_2020}                           & 200                               & {\color[HTML]{FF0000} \textbf{0.1336}} & 0.0201                                 \\
GraphCSPN\cite{liu_graphcspn_2022}                       & 200                               & 0.1338                                 & 0.0197                                 \\
CompletionFormer\cite{youmin_completionformer_2023}                & 200                               & 0.1349                                 & 0.0205                                 \\
\textbf{Ours}        & 200          & 0.1346                                 & {\color[HTML]{FF0000} \textbf{0.0191}} \\ \midrule
NLSPN\cite{vedaldi_non-local_2020}                           & 50                               & 0.2123                                 & 0.0382                                 \\
GraphCSPN\cite{liu_graphcspn_2022}                       & 50                               & 0.2119                                 & 0.0375                                 \\
CompletionFormer\cite{youmin_completionformer_2023}                & 50                               & 0.2183                                 & 0.0405                                 \\
\textbf{Ours}        & 50           & {\color[HTML]{FF0000} \textbf{0.2079}} & {\color[HTML]{FF0000} \textbf{0.0350}} \\ \midrule
NLSPN\cite{vedaldi_non-local_2020}                           & 10                               & 0.3639                                 & 0.0827                                 \\
GraphCSPN\cite{liu_graphcspn_2022}                       & 10                               & 0.3650                                 & 0.0832                                 \\
CompletionFormer\cite{youmin_completionformer_2023}                & 10                               & 0.3642                                 & 0.0874                                 \\
\textbf{Ours}        & 10           & {\color[HTML]{FF0000} \textbf{0.3226}} & {\color[HTML]{FF0000} \textbf{0.0681}} \\ \midrule
NLSPN\cite{vedaldi_non-local_2020}                           & 5                               & 0.4439                                 & 0.1094                                 \\
GraphCSPN\cite{liu_graphcspn_2022}                       & 5                               & 0.4555                                 & 0.1151                                 \\
CompletionFormer\cite{youmin_completionformer_2023}                & 5                               & 0.4377                                 & 0.1144                                 \\
\textbf{Ours}        & 5            & {\color[HTML]{FF0000} \textbf{0.3816}} & {\color[HTML]{FF0000} \textbf{0.0877}} \\ \midrule
NLSPN\cite{vedaldi_non-local_2020}                           & shift\_grid                               & 0.4263                                 & 0.1043                                 \\
GraphCSPN\cite{liu_graphcspn_2022}                       & shift\_grid                               & 0.4485                                 & 0.1083                                 \\
CompletionFormer\cite{youmin_completionformer_2023}                & shift\_grid                               & 0.4313                                 & 0.1149                                 \\
\textbf{Ours} & shift\_grid  & {\color[HTML]{FF0000} \textbf{0.3736}} & {\color[HTML]{FF0000} \textbf{0.0852}} \\ \midrule
NLSPN\cite{vedaldi_non-local_2020}                           & uneven density                               & 0.3931                                & 0.0909                                 \\
GraphCSPN\cite{liu_graphcspn_2022}                       & uneven density                               & 0.3991                                 & 0.0920                                 \\
CompletionFormer\cite{youmin_completionformer_2023}                & uneven density                               & 0.3919                                 & 0.0961                                 \\
\textbf{Ours} & uneven density & {\color[HTML]{FF0000} \textbf{0.3543}} & {\color[HTML]{FF0000} \textbf{0.0763}} \\ \midrule
NLSPN\cite{vedaldi_non-local_2020}                           & ORB keypoint                               & 0.2590                                 & 0.0625                                 \\
GraphCSPN\cite{liu_graphcspn_2022}                       & ORB keypoint                               & 0.2571                                 & 0.0603                                 \\
CompletionFormer\cite{youmin_completionformer_2023}                & ORB keypoint                               & 0.2660                                 & 0.0672                                 \\
\textbf{Ours} & ORB keypoint & {\color[HTML]{FF0000} \textbf{0.2441}} & {\color[HTML]{FF0000} \textbf{0.0541}} \\ \midrule
NLSPN\cite{vedaldi_non-local_2020}                           & big holes                               & 0.2368                                 & 0.0402                                 \\
GraphCSPN\cite{liu_graphcspn_2022}                       & big holes                               & 0.2443                                 & 0.0403                                 \\
CompletionFormer\cite{youmin_completionformer_2023}                & big holes                               & 0.2493                                 & 0.0434                                 \\
\textbf{Ours} & big holes    & {\color[HTML]{FF0000} \textbf{0.2262}} & {\color[HTML]{FF0000} \textbf{0.0362}} \\ \bottomrule
\end{tabular}
\caption{\textbf{Results on NYU Depth V2 with different patterns.} Quantitative comparisons with state-of-the-art methods in different spatial patterns. The best results are red and in bold.}
\vspace{-2em}
\label{tab:nyu}
\end{table}

We designed the following scenarios during testing to simulate varying sparsity and non-uniform inputs as shown in \cref{patterns}. 1) Varying sparsity inputs: Randomly sampling depth points included some extreme cases, such as only 5 valid depth points. 2) VCSEL TOF sensors\cite{luetzenburg2021evaluation}: Sampling a shifting grid from the ground truth. 3) Uneven density: Randomly selected two regions with different sizes and different numbers of depth points from the ground truth. 4) Keypoint-based Sampling: Following the SpaseSPN\cite{sparsespn}, we use a keypoints-based(ORB\cite{rublee2011orb}) sample strategy to simulate the sparse depth values available from keypoint triangulation. 5) Holes: Randomly select depth points, then select a $200 \times 200$ square grid, and remove all depth points inside the grid.

The quantitative comparison results on NYU Depth V2 under various non-uniform settings are shown in \cref{tab:nyu} and some visual comparisons are given in \cref{fig:nyu}. It can be seen that SparseDC consistently outperforms the baselines only except the 200-sample case. Moreover, our method achieved 17\% improvements on REL and 7.8\% improvements on RMSE, respectively, over the State-of-the-art method CompletionFormer\cite{youmin_completionformer_2023}. This is mainly because SparseDC can dynamically adjust the reliance on both local and global features during the depth completion process and thus can achieve a balance to retain the
precise geometry of regions with available depth values and accurate structures in regions with no depth input.

\begin{table}[t]
    \centering
    \footnotesize
    \begin{tabular}{ccccccc}
        \toprule
        Methods                                                & Params & RMSE(m)$\downarrow$                    & REL$\downarrow$                        \\ \midrule
        NLSPN\cite{vedaldi_non-local_2020}                     & 25.8M  & 0.5966                                 & 0.1257                                 \\
        NLSPN*\cite{vedaldi_non-local_2020}                    & 25.8M  & 3.5966                                 & 1.9789                                 \\
        GraphCSPN$*$\cite{liu_graphcspn_2022}                  & 26M    & 4.3420                                 & 2.2700                                 \\
        CompletionFormer\cite{youmin_completionformer_2023}    & 45M    & 0.9613                                 & 0.4707                                 \\
        CompletionFormer$*$\cite{youmin_completionformer_2023} & 45M    & 0.6047                                 & 0.1574                                 \\ \midrule
        \textbf{Ours}                                          & 38.2M  & {\color[HTML]{FF0000} \textbf{0.4536}} & {\color[HTML]{FF0000} \textbf{0.1180}} \\  \bottomrule
    \end{tabular}
    \caption{\textbf{Generalization Ability Test on SUN RGBD.} $*$ denotes trained model using our training strategy, without $*$ is their released pretrained model trained using fixed patterns.}
    \vspace{-1em}
    \label{tab:sun}
\end{table}
\subsection{Generalization results on SUN RGB-D}
To validate the generalization ability of our method, we use the model trained on NYU\cite{nyu} to conduct depth completion on SUN RGB-D directly. In the test set, we used the noisy and incomplete raw depth image as the input without sampling. As shown in \cref{tab:sun}, our proposed SparseDC performs best in most metrics. The augment-based strategy makes \cite{vedaldi_non-local_2020,liu_graphcspn_2022} overfit the training data. Completionformer\cite{youmin_completionformer_2023} and SparseDC fuse global and local information simultaneously, resulting in improvements and robustness with unseen patterns. We report the corresponding visual comparison results in the Appendix.

\subsection{Results on KITTI Depth Completion}
On the outdoor KITTI dataset, we randomly mask some rows in depth maps during training as mentioned in \cref{sec:dataset}. Due to resource constraints, we directly use the metric in reported CompltionFormer\cite{youmin_completionformer_2023}
The specific results are shown in \cref{tab:kitti}. It can be seen that SparseDC consistently outperforms the baselines by a large margin in MAE, and the gap becomes larger when fewer scanning lines are kept, demonstrating the effectiveness of SparseDC under sparse input.

\subsection{Ablation Study}
To demonstrate the effectiveness of each component proposed in SparseDC, we conducted corresponding ablation experiments on NYU Depth V2 dataset by removing each component as follows.
\begin{table}[t]
    \centering
    \footnotesize
    \begin{tabular}{cccc}
    \toprule
Methods          & Scanning Lines & RMSE(m)$\downarrow$                                & MAE(m)$\downarrow$                                 \\ \midrule
NLSPN\cite{vedaldi_non-local_2020}            & lines4        & 2.2648                                 & 0.7492                                 \\
PENet\cite{hu_penet_2021}            & lines4        & 2.2323                                 & 0.8132 \\
CompletionFormer\cite{youmin_completionformer_2023} & lines4        & 2.1500                                 & 0.7401 \\
DySPN\cite{lin_dynamic_2022}            & lines4        & 2.2858 & 0.8343 \\
\textbf{Ours}             & lines4        & {\color[HTML]{FF0000} \textbf{2.1029}} & {\color[HTML]{FF0000} \textbf{0.7012}} \\ \midrule
NLSPN\cite{vedaldi_non-local_2020}            & lines8        & 1.5452                                 & 0.4593 \\
PENet\cite{hu_penet_2021}            & lines8        & 1.5983                                 & 0.5261                                 \\
\textbf{Ours}             & lines8        & {\color[HTML]{FF0000} \textbf{1.5286}} & {\color[HTML]{FF0000} \textbf{0.4444}} \\ \midrule
NLSPN\cite{vedaldi_non-local_2020}            & lines16       & 1.1781                                 & 0.3274                                 \\
PENet\cite{hu_penet_2021}            & lines16       & 1.2231                                 & 0.3818                                 \\
CompletionFormer\cite{youmin_completionformer_2023} & lines16       & 1.2186                                 & 0.3374                                 \\
DySPN\cite{lin_dynamic_2022}            & lines16       & 1.2748 & 0.3664 \\
\textbf{Ours}             & lines16       & {\color[HTML]{FF0000} \textbf{1.1764}} & {\color[HTML]{FF0000} \textbf{0.3178}} \\ \midrule
NLSPN\cite{vedaldi_non-local_2020}            & lines32       & 0.9575                                 & 0.2587                                 \\
PENet\cite{hu_penet_2021}            & lines32       & 0.9866                                 & 0.3019                                 \\
\textbf{Ours}             & lines32       & {\color[HTML]{FF0000} \textbf{0.9485}} & {\color[HTML]{FF0000} \textbf{0.2488}} \\ \midrule
NLSPN\cite{vedaldi_non-local_2020}            & lines64       & 0.8125                                 & 0.2167                                 \\
PENet\cite{hu_penet_2021}            & lines64       & 0.8406                                 & 0.2472                                 \\
CompletionFormer\cite{youmin_completionformer_2023} & lines64       & 0.8487                                 & 0.2159                                 \\
DySPN\cite{lin_dynamic_2022}            & lines64       & 0.8785 & 0.2286 \\
\textbf{Ours}             & lines64       & {\color[HTML]{FF0000} \textbf{0.7966}} & {\color[HTML]{FF0000} \textbf{0.2051}} \\
\bottomrule
\end{tabular}
\caption{\textbf{Results on KITTI DC.} Quantitative comparisons with state-of-the-art methods on varying scanning lines. }
\vspace{-1em}
\label{tab:kitti}
\end{table}
\begin{figure}[t]
    \centering
    \includegraphics[width=1.0\linewidth]{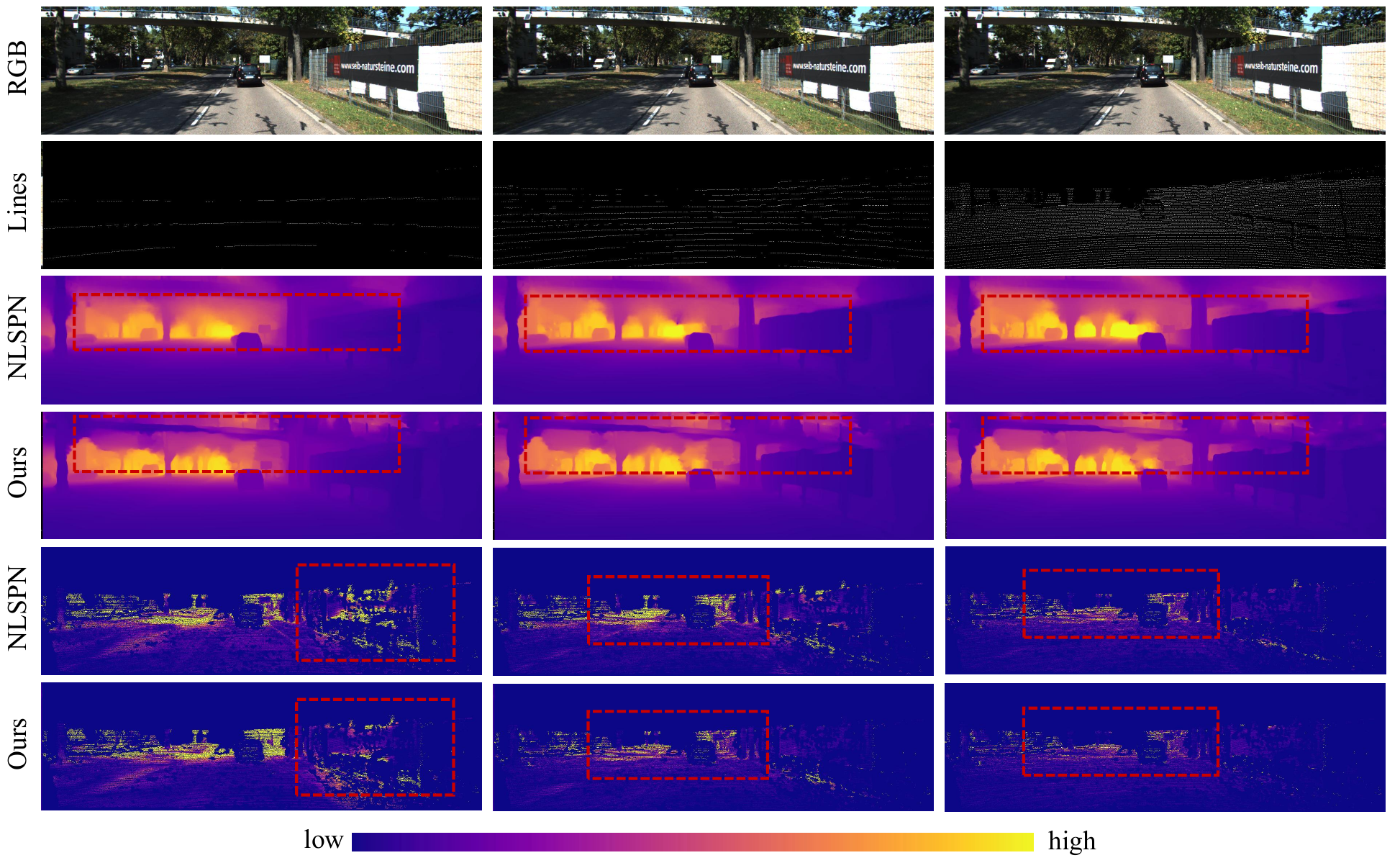}
    \caption{\textbf{Qualitative results on KITTI DC.} We report the completed depth and error map with varying scanning lines as inputs.}
    \vspace{-1em}
    \label{fig:kitti}
\end{figure}
\begin{itemize}
    \item Remove the Sparse Feature Filling Module, and replace it with two simple convolution layers, as previous methods\cite{vedaldi_non-local_2020,liu_graphcspn_2022}.
    \item Remove the Uncertainty-Guide Feature Fusion Module, and use \cref{eq:fuse1} to fuse the extracted features.
\begin{equation}\label{eq:fuse1}
    f^n = Gate(f^n_{l} \oplus \cdot f^n_{g})
\end{equation}
    \item Removing the two-branch feature extraction structure in favor of a single-branch one, i.e., the overall model will degenerate into a combination of ResNet\cite{resnet} with a non-local spatial propagation module, so we use the retrained NLSPN\cite{vedaldi_non-local_2020} as a replacement.
\end{itemize}
\begin{table}[t]
    \centering
    \footnotesize
    \begin{tabular}{cccc}
        \toprule
        Model               & samples        & RMSE(m)$\downarrow$                    & REL$\downarrow$                        \\ \midrule
        w/o UFFM            & 500            & {\color[HTML]{FF0000} \textbf{0.0973}} & {\color[HTML]{FF0000} \textbf{0.0128}} \\
        w/o SFFM            & 500            & 0.0974                                 & {\color[HTML]{FF0000} \textbf{0.0128}} \\
        w/o Two-Branch      & 500            & 0.0984                                 & 0.0136                                 \\
        \textbf{Full Model} & 500            & 0.0976                                 & {\color[HTML]{FF0000} \textbf{0.0128}} \\ \midrule
        w/o UFFM            & 50             & 0.2089                                 & 0.0356                                 \\
        w/o SFFM            & 50             & 0.2092                                 & 0.0354                                 \\
        w/o Two-Branch      & 50             & 0.2123                                 & 0.0382                                 \\
        \textbf{Full Model} & 50             & {\color[HTML]{FF0000} \textbf{0.2079}} & {\color[HTML]{FF0000} \textbf{0.0350}} \\ \midrule
        w/o UFFM            & 5              & 0.3846                                 & 0.0904                                 \\
        w/o SFFM            & 5              & 0.3854                                 & 0.0918                                 \\
        w/o Two-Branch      & 5              & 0.4439                                 & 0.1094                                 \\
        \textbf{Full Model} & 5              & {\color[HTML]{FF0000} \textbf{0.3816}} & {\color[HTML]{FF0000} \textbf{0.0877}} \\ \midrule
        w/o UFFM            & shift\_grid    & 0.4618                                 & 0.1072                                 \\
        w/o SFFM            & shift\_grid    & 0.4380                                 & 0.1042                                 \\
        w/o Two-Branch      & shift\_grid    & 0.4263                                 & 0.1043                                 \\
        \textbf{Full Model} & shift\_grid    & {\color[HTML]{FF0000} \textbf{0.3736}} & {\color[HTML]{FF0000} \textbf{0.0852}} \\ \midrule
        w/o UFFM            & uneven density & 0.5391                                 & 0.1266                                 \\
        w/o SFFM            & uneven density & 0.4124                                 & 0.0934                                 \\
        w/o Two-Branch      & uneven density & 0.3931                                 & 0.0909                                 \\
        \textbf{Full Model} & uneven density & {\color[HTML]{FF0000} \textbf{0.3543}} & {\color[HTML]{FF0000} \textbf{0.0763}} \\
        \bottomrule
    \end{tabular}
    \caption{\textbf{Ablation study.} The evaluation results of different components of SparseDC.}
    \vspace{-1em}
    \label{tab:ablation}
\end{table}
All ablation models are trained under the same settings as SparseDC, the results are shown in \cref{tab:ablation}.

Our full pipeline gets the minimum RMSE and REL value for most experiments, removing any component will degrade the overall performance of the network. In addition, the results show that removing the Two-Branch feature extraction structure has the greatest impact on the network in the face of extremely sparse inputs. This is because the Two-Branch structure provides both global semantic and local geometric information, which is necessary for dealing with non-uniform inputs. Furthermore,  SFFM provides a stabilized feature space for sparse depth maps, which makes the depth estimated by the network more accurate. UFFM can predict the uncertainty based on the spatial pattern of the inputs, and then explicitly guides the fusion process of features from different branches by this uncertainty, as shown in \cref{fig:weight}. Both modules benefit SparseDC with sparse and non-uniform input.
\section{Conclusion}
\label{sec:conclu}
This paper focused on depth completion of \textit{sparse} and \textit{non-uniform} inputs. To address this, We first propose an SFFM module to improve the feature stability by explicitly filling the unstable depth features with stable image features. Then, we introduce a two-branch feature embedder to extract local or long-term information of the input depth map and RGB image by CNNs and ViTs. Here we propose an uncertainty-based fusion module called UFFM to predict both precise local geometries of regions with available depth values and accurate structures in regions with no depth input. Extensive indoor and outdoor experiments demonstrate our method's effectiveness when facing sparse and non-uniform inputs.


{
    \small
    \bibliographystyle{ieeenat_fullname}
    \bibliography{main}

\begin{thebibliography}{58}
\providecommand{\natexlab}[1]{#1}
\providecommand{\url}[1]{\texttt{#1}}
\expandafter\ifx\csname urlstyle\endcsname\relax
  \providecommand{\doi}[1]{doi: #1}\else
  \providecommand{\doi}{doi: \begingroup \urlstyle{rm}\Url}\fi

\bibitem[Agarap(2018)]{relu}
Abien~Fred Agarap.
\newblock Deep learning using rectified linear units (relu).
\newblock \emph{arXiv preprint arXiv:1803.08375}, 2018.

\bibitem[Atanacio-Jim{\'e}nez et~al.(2011)Atanacio-Jim{\'e}nez,
  Gonz{\'a}lez-Barbosa, Hurtado-Ramos, Ornelas-Rodr{\'\i}guez,
  Jim{\'e}nez-Hern{\'a}ndez, Garc{\'\i}a-Ramirez, and
  Gonz{\'a}lez-Barbosa]{atanacio2011lidar}
Gerardo Atanacio-Jim{\'e}nez, Jos{\'e}-Joel Gonz{\'a}lez-Barbosa, Juan~B
  Hurtado-Ramos, Francisco~J Ornelas-Rodr{\'\i}guez, Hugo
  Jim{\'e}nez-Hern{\'a}ndez, Teresa Garc{\'\i}a-Ramirez, and Ricardo
  Gonz{\'a}lez-Barbosa.
\newblock Lidar velodyne hdl-64e calibration using pattern planes.
\newblock \emph{International Journal of Advanced Robotic Systems}, 8\penalty0
  (5):\penalty0 59, 2011.

\bibitem[Cheng et~al.(2018)Cheng, Wang, and Yang]{ferrari_depth_2018}
Xinjing Cheng, Peng Wang, and Ruigang Yang.
\newblock Depth {Estimation} via {Affinity} {Learned} with {Convolutional}
  {Spatial} {Propagation} {Network}.
\newblock In \emph{Computer {Vision} – {ECCV} 2018}, pages 108--125. Springer
  International Publishing, Cham, 2018.

\bibitem[Cheng et~al.(2019)Cheng, Wang, Guan, and Yang]{cheng_cspn_2019}
Xinjing Cheng, Peng Wang, Chenye Guan, and Ruigang Yang.
\newblock {CSPN}++: {Learning} {Context} and {Resource} {Aware} {Convolutional}
  {Spatial} {Propagation} {Networks} for {Depth} {Completion}, 2019.

\bibitem[Conti et~al.(2023)Conti, Poggi, and Mattoccia]{SpAgnet}
Andrea Conti, Matteo Poggi, and Stefano Mattoccia.
\newblock Sparsity agnostic depth completion.
\newblock In \emph{Proceedings of the IEEE/CVF Winter Conference on
  Applications of Computer Vision}, pages 5871--5880, 2023.

\bibitem[Contributors(2022)]{spconv2022}
Spconv Contributors.
\newblock Spconv: Spatially sparse convolution library.
\newblock \url{https://github.com/traveller59/spconv}, 2022.

\bibitem[Cordts et~al.(2016)Cordts, Omran, Ramos, Rehfeld, Enzweiler, Benenson,
  Franke, Roth, and Schiele]{cordts2016cityscapes}
Marius Cordts, Mohamed Omran, Sebastian Ramos, Timo Rehfeld, Markus Enzweiler,
  Rodrigo Benenson, Uwe Franke, Stefan Roth, and Bernt Schiele.
\newblock The cityscapes dataset for semantic urban scene understanding.
\newblock In \emph{Proceedings of the IEEE conference on computer vision and
  pattern recognition}, pages 3213--3223, 2016.

\bibitem[Geiger et~al.(2011)Geiger, Ziegler, and Stiller]{geiger2011stereoscan}
Andreas Geiger, Julius Ziegler, and Christoph Stiller.
\newblock Stereoscan: Dense 3d reconstruction in real-time.
\newblock In \emph{2011 IEEE intelligent vehicles symposium (IV)}, pages
  963--968. Ieee, 2011.

\bibitem[Geiger et~al.(2012)Geiger, Lenz, and Urtasun]{kitti}
Andreas Geiger, Philip Lenz, and Raquel Urtasun.
\newblock Are we ready for autonomous driving? the kitti vision benchmark
  suite.
\newblock In \emph{Conference on Computer Vision and Pattern Recognition
  (CVPR)}, 2012.

\bibitem[Graham and Van~der Maaten(2017)]{graham2017submanifold}
Benjamin Graham and Laurens Van~der Maaten.
\newblock Submanifold sparse convolutional networks.
\newblock \emph{arXiv preprint arXiv:1706.01307}, 2017.

\bibitem[Gu et~al.(2021)Gu, Xiang, Ye, and Wang]{gu_denselidar_2021}
Jiaqi Gu, Zhiyu Xiang, Yuwen Ye, and Lingxuan Wang.
\newblock {DenseLiDAR}: {A} {Real}-{Time} {Pseudo} {Dense} {Depth} {Guided}
  {Depth} {Completion} {Network}.
\newblock \emph{IEEE Robotics and Automation Letters}, 6\penalty0 (2):\penalty0
  1808--1815, 2021.

\bibitem[He et~al.(2016)He, Zhang, Ren, and Sun]{resnet}
Kaiming He, Xiangyu Zhang, Shaoqing Ren, and Jian Sun.
\newblock Deep residual learning for image recognition.
\newblock In \emph{Proceedings of the IEEE conference on computer vision and
  pattern recognition}, pages 770--778, 2016.

\bibitem[Hu et~al.(2019)Hu, Ozay, Zhang, and Okatani]{hu2019revisiting}
Junjie Hu, Mete Ozay, Yan Zhang, and Takayuki Okatani.
\newblock Revisiting single image depth estimation: Toward higher resolution
  maps with accurate object boundaries.
\newblock In \emph{2019 IEEE winter conference on applications of computer
  vision (WACV)}, pages 1043--1051. IEEE, 2019.

\bibitem[Hu et~al.(2023)Hu, Bao, Ozay, Fan, Gao, Liu, and Lam]{hu_deep_2023}
Junjie Hu, Chenyu Bao, Mete Ozay, Chenyou Fan, Qing Gao, Honghai Liu, and
  Tin~Lun Lam.
\newblock Deep {Depth} {Completion} from {Extremely} {Sparse} {Data}: {A}
  {Survey}.
\newblock \emph{IEEE Transactions on Pattern Analysis and Machine
  Intelligence}, 2023.

\bibitem[Hu et~al.(2021)Hu, Wang, Li, Ning, Fan, and Gong]{hu_penet_2021}
Mu Hu, Shuling Wang, Bin Li, Shiyu Ning, Li Fan, and Xiaojin Gong.
\newblock {PENet}: {Towards} {Precise} and {Efficient} {Image} {Guided} {Depth}
  {Completion}, 2021.

\bibitem[Ioffe and Szegedy(2015)]{batchnorm}
Sergey Ioffe and Christian Szegedy.
\newblock Batch normalization: Accelerating deep network training by reducing
  internal covariate shift.
\newblock In \emph{International conference on machine learning}, pages
  448--456. pmlr, 2015.

\bibitem[Jeon et~al.(2022)Jeon, Kim, and Seo]{jeon_abcd_2022}
Yurim Jeon, Hwichang Kim, and Seung-Woo Seo.
\newblock {ABCD}: {Attentive} {Bilateral} {Convolutional} {Network} for
  {Robust} {Depth} {Completion}.
\newblock \emph{IEEE Robotics and Automation Letters}, 7\penalty0 (1):\penalty0
  81--87, 2022.

\bibitem[Jiang et~al.(2021)Jiang, Ding, Hu, and Huang]{jiang2021plnet}
Hualie Jiang, Laiyan Ding, Junjie Hu, and Rui Huang.
\newblock Plnet: Plane and line priors for unsupervised indoor depth
  estimation.
\newblock In \emph{2021 International Conference on 3D Vision (3DV)}, pages
  741--750. IEEE, 2021.

\bibitem[Keselman et~al.(2017)Keselman, Iselin~Woodfill, Grunnet-Jepsen, and
  Bhowmik]{keselman2017intel}
Leonid Keselman, John Iselin~Woodfill, Anders Grunnet-Jepsen, and Achintya
  Bhowmik.
\newblock Intel realsense stereoscopic depth cameras.
\newblock In \emph{Proceedings of the IEEE conference on computer vision and
  pattern recognition workshops}, pages 1--10, 2017.

\bibitem[Krizhevsky et~al.(2012)Krizhevsky, Sutskever, and Hinton]{imagenet}
Alex Krizhevsky, Ilya Sutskever, and Geoffrey~E Hinton.
\newblock Imagenet classification with deep convolutional neural networks.
\newblock \emph{Advances in neural information processing systems}, 25, 2012.

\bibitem[Li et~al.(2009)Li, Socher, and Fei-Fei]{li2009towards}
Li-Jia Li, Richard Socher, and Li Fei-Fei.
\newblock Towards total scene understanding: Classification, annotation and
  segmentation in an automatic framework.
\newblock In \emph{2009 IEEE Conference on Computer Vision and Pattern
  Recognition}, pages 2036--2043. IEEE, 2009.

\bibitem[Lin et~al.(2022)Lin, Cheng, Zhong, Zhou, and Yang]{lin_dynamic_2022}
Yuankai Lin, Tao Cheng, Qi Zhong, Wending Zhou, and Hua Yang.
\newblock Dynamic {Spatial} {Propagation} {Network} for {Depth} {Completion}.
\newblock \emph{Proceedings of the AAAI Conference on Artificial Intelligence},
  36\penalty0 (2):\penalty0 1638--1646, 2022.

\bibitem[Liu et~al.(2020)Liu, Song, Lyu, Diao, Wang, Liu, and
  Zhang]{liu_fcfr-net_2020}
Lina Liu, Xibin Song, Xiaoyang Lyu, Junwei Diao, Mengmeng Wang, Yong Liu, and
  Liangjun Zhang.
\newblock {FCFR}-{Net}: {Feature} {Fusion} based {Coarse}-to-{Fine} {Residual}
  {Learning} for {Depth} {Completion}.
\newblock page~9, 2020.

\bibitem[Liu et~al.(2021)Liu, Liao, Wang, Geiger, and Liu]{liu_learning_2021}
Lina Liu, Yiyi Liao, Yue Wang, Andreas Geiger, and Yong Liu.
\newblock Learning {Steering} {Kernels} for {Guided} {Depth} {Completion}.
\newblock \emph{IEEE Transactions on Image Processing}, 30:\penalty0
  2850--2861, 2021.

\bibitem[Liu et~al.(2022)Liu, Shao, Wang, Li, and Wang]{liu_graphcspn_2022}
Xin Liu, Xiaofei Shao, Bo Wang, Yali Li, and Shengjin Wang.
\newblock {GraphCSPN}: {Geometry}-{Aware} {Depth} {Completion} via {Dynamic}
  {GCNs}, 2022.

\bibitem[Long et~al.(2022)Long, Zhang, Li, Wang, Dong, and Yang]{long2022pc2}
Chen Long, WenXiao Zhang, Ruihui Li, Hao Wang, Zhen Dong, and Bisheng Yang.
\newblock Pc2-pu: Patch correlation and point correlation for effective point
  cloud upsampling.
\newblock In \emph{Proceedings of the 30th ACM International Conference on
  Multimedia}, pages 2191--2201, 2022.

\bibitem[Luetzenburg et~al.(2021)Luetzenburg, Kroon, and
  Bj{\o}rk]{luetzenburg2021evaluation}
Gregor Luetzenburg, Aart Kroon, and Anders~A Bj{\o}rk.
\newblock Evaluation of the apple iphone 12 pro lidar for an application in
  geosciences.
\newblock \emph{Scientific reports}, 11\penalty0 (1):\penalty0 22221, 2021.

\bibitem[Ma and Karaman(2018)]{sparse2dense}
Fangchang Ma and Sertac Karaman.
\newblock Sparse-to-dense: Depth prediction from sparse depth samples and a
  single image.
\newblock In \emph{2018 IEEE international conference on robotics and
  automation (ICRA)}, pages 4796--4803. IEEE, 2018.

\bibitem[Maurer et~al.(2016)Maurer, Gerdes, Lenz, and
  Winner]{maurer2016autonomous}
Markus Maurer, J~Christian Gerdes, Barbara Lenz, and Hermann Winner.
\newblock \emph{Autonomous driving: technical, legal and social aspects}.
\newblock Springer Nature, 2016.

\bibitem[Metzger et~al.(2022)Metzger, Daudt, and
  Schindler]{metzger_guided_2022}
Nando Metzger, Rodrigo~Caye Daudt, and Konrad Schindler.
\newblock Guided {Depth} {Super}-{Resolution} by {Deep} {Anisotropic}
  {Diffusion}, 2022.

\bibitem[Märkert et~al.(2022)Märkert, Sunkel, Haselhoff, and
  Rudolph]{markert_segmentation-guided_2022}
Fabian Märkert, Martin Sunkel, Anselm Haselhoff, and Stefan Rudolph.
\newblock Segmentation-guided {Domain} {Adaptation} for {Efficient} {Depth}
  {Completion}, 2022.

\bibitem[Nathan~Silberman and Fergus(2012)]{nyu}
Pushmeet~Kohli Nathan~Silberman, Derek~Hoiem and Rob Fergus.
\newblock Indoor segmentation and support inference from rgbd images.
\newblock In \emph{ECCV}, 2012.

\bibitem[Pan et~al.(2016)Pan, Guan, Luo, Duan, Tian, Yi, Zhao, and
  Yu]{pan2016dense}
Hailong Pan, Tao Guan, Yawei Luo, Liya Duan, Yuan Tian, Liu Yi, Yizhu Zhao, and
  Junqing Yu.
\newblock Dense 3d reconstruction combining depth and rgb information.
\newblock \emph{Neurocomputing}, 175:\penalty0 644--651, 2016.

\bibitem[Park et~al.(2020)Park, Joo, Hu, Liu, and
  So~Kweon]{vedaldi_non-local_2020}
Jinsun Park, Kyungdon Joo, Zhe Hu, Chi-Kuei Liu, and In So~Kweon.
\newblock Non-local {Spatial} {Propagation} {Network} for {Depth} {Completion}.
\newblock In \emph{Computer {Vision} – {ECCV} 2020}, pages 120--136. Springer
  International Publishing, Cham, 2020.

\bibitem[Park et~al.(2019)Park, Kim, Lee, Na, and Jeon]{park2019literature}
Mi~Jin Park, Dong~Jun Kim, Unjoo Lee, Eun~Jin Na, and Hong~Jin Jeon.
\newblock A literature overview of virtual reality (vr) in treatment of
  psychiatric disorders: recent advances and limitations.
\newblock \emph{Frontiers in psychiatry}, 10:\penalty0 505, 2019.

\bibitem[Qiu et~al.(2019)Qiu, Cui, Zhang, Zhang, Liu, Zeng, and
  Pollefeys]{deeplidar}
Jiaxiong Qiu, Zhaopeng Cui, Yinda Zhang, Xingdi Zhang, Shuaicheng Liu, Bing
  Zeng, and Marc Pollefeys.
\newblock Deeplidar: Deep surface normal guided depth prediction for outdoor
  scene from sparse lidar data and single color image.
\newblock In \emph{Proceedings of the IEEE/CVF Conference on Computer Vision
  and Pattern Recognition}, pages 3313--3322, 2019.

\bibitem[Rho et~al.(2020)Rho, Ha, and Kim]{rho_guideformer_2020}
Kyeongha Rho, Jinsung Ha, and Youngjung Kim.
\newblock {GuideFormer}: {Transformers} for {Image} {Guided} {Depth}
  {Completion}.
\newblock page~10, 2020.

\bibitem[Ronneberger et~al.(2015)Ronneberger, Fischer, and Brox]{unet}
Olaf Ronneberger, Philipp Fischer, and Thomas Brox.
\newblock U-net: Convolutional networks for biomedical image segmentation.
\newblock In \emph{Medical Image Computing and Computer-Assisted
  Intervention--MICCAI 2015: 18th International Conference, Munich, Germany,
  October 5-9, 2015, Proceedings, Part III 18}, pages 234--241. Springer, 2015.

\bibitem[Rublee et~al.(2011)Rublee, Rabaud, Konolige, and
  Bradski]{rublee2011orb}
Ethan Rublee, Vincent Rabaud, Kurt Konolige, and Gary Bradski.
\newblock Orb: An efficient alternative to sift or surf.
\newblock In \emph{2011 International conference on computer vision}, pages
  2564--2571. Ieee, 2011.

\bibitem[Shao et~al.(2023)Shao, Pei, Chen, Li, Liu, and Li]{shao2023urcdc}
Shuwei Shao, Zhongcai Pei, Weihai Chen, Ran Li, Zhong Liu, and Zhengguo Li.
\newblock Urcdc-depth: Uncertainty rectified cross-distillation with cutflip
  for monocular depth estimation.
\newblock \emph{arXiv preprint arXiv:2302.08149}, 2023.

\bibitem[Song et~al.(2015)Song, Lichtenberg, and Xiao]{song2015sun}
Shuran Song, Samuel~P Lichtenberg, and Jianxiong Xiao.
\newblock Sun rgb-d: A rgb-d scene understanding benchmark suite.
\newblock In \emph{Proceedings of the IEEE conference on computer vision and
  pattern recognition}, pages 567--576, 2015.

\bibitem[Tang et~al.(2019)Tang, Tian, Feng, Li, and Tan]{tang_learning_2019}
Jie Tang, Fei-Peng Tian, Wei Feng, Jian Li, and Ping Tan.
\newblock Learning {Guided} {Convolutional} {Network} for {Depth} {Completion},
  2019.

\bibitem[Uhrig et~al.(2017)Uhrig, Schneider, Schneider, Franke, Brox, and
  Geiger]{Sparsitycnn}
Jonas Uhrig, Nick Schneider, Lukas Schneider, Uwe Franke, Thomas Brox, and
  Andreas Geiger.
\newblock Sparsity invariant cnns.
\newblock In \emph{2017 international conference on 3D Vision (3DV)}, pages
  11--20. IEEE, 2017.

\bibitem[Wang et~al.(2021)Wang, Xie, Li, Fan, Song, Liang, Lu, Luo, and
  Shao]{pvt}
Wenhai Wang, Enze Xie, Xiang Li, Deng-Ping Fan, Kaitao Song, Ding Liang, Tong
  Lu, Ping Luo, and Ling Shao.
\newblock Pyramid vision transformer: A versatile backbone for dense prediction
  without convolutions.
\newblock In \emph{Proceedings of the IEEE/CVF international conference on
  computer vision}, pages 568--578, 2021.

\bibitem[Wang et~al.(2022)Wang, Xie, Li, Fan, Song, Liang, Lu, Luo, and
  Shao]{pvtv2}
Wenhai Wang, Enze Xie, Xiang Li, Deng-Ping Fan, Kaitao Song, Ding Liang, Tong
  Lu, Ping Luo, and Ling Shao.
\newblock Pvtv2: Improved baselines with pyramid vision transformer.
\newblock \emph{Computational Visual Media}, 8\penalty0 (3):\penalty0 1--10,
  2022.

\bibitem[Wang et~al.(2023)Wang, Li, Zhang, Liu, Gao, and Dai]{wang2023lrru}
Yufei Wang, Bo Li, Ge Zhang, Qi Liu, Tao Gao, and Yuchao Dai.
\newblock Lrru: Long-short range recurrent updating networks for depth
  completion.
\newblock In \emph{Proceedings of the IEEE/CVF International Conference on
  Computer Vision}, pages 9422--9432, 2023.

\bibitem[Woo et~al.(2023)Woo, Debnath, Hu, Chen, Liu, Kweon, and
  Xie]{woo2023convnext}
Sanghyun Woo, Shoubhik Debnath, Ronghang Hu, Xinlei Chen, Zhuang Liu, In~So
  Kweon, and Saining Xie.
\newblock Convnext v2: Co-designing and scaling convnets with masked
  autoencoders.
\newblock In \emph{Proceedings of the IEEE/CVF Conference on Computer Vision
  and Pattern Recognition}, pages 16133--16142, 2023.

\bibitem[Wu et~al.(2022)Wu, Lee, and Hoiem]{sparsespn}
Yuqun Wu, Jae~Yong Lee, and Derek Hoiem.
\newblock Sparse spn: Depth completion from sparse keypoints.
\newblock \emph{arXiv preprint arXiv:2212.00987}, 2022.

\bibitem[Xu et~al.(2020)Xu, Li, Du, Zhang, and Liu]{leakyrelu}
Jin Xu, Zishan Li, Bowen Du, Miaomiao Zhang, and Jing Liu.
\newblock Reluplex made more practical: Leaky relu.
\newblock In \emph{2020 IEEE Symposium on Computers and communications (ISCC)},
  pages 1--7. IEEE, 2020.

\bibitem[Xu et~al.(2019)Xu, Zhu, Shi, Zhang, Bao, and Li]{xu2019depth}
Yan Xu, Xinge Zhu, Jianping Shi, Guofeng Zhang, Hujun Bao, and Hongsheng Li.
\newblock Depth completion from sparse lidar data with depth-normal
  constraints.
\newblock In \emph{Proceedings of the IEEE/CVF International Conference on
  Computer Vision}, pages 2811--2820, 2019.

\bibitem[Yan et~al.(2022)Yan, Wang, Li, Zhang, Li, and Yang]{yan_rignet_2022}
Zhiqiang Yan, Kun Wang, Xiang Li, Zhenyu Zhang, Jun Li, and Jian Yang.
\newblock {RigNet}: {Repetitive} {Image} {Guided} {Network} for {Depth}
  {Completion}, 2022.

\bibitem[Youmin et~al.(2023)Youmin, Xianda, Matteo, Zheng, Guan, and
  Stefano]{youmin_completionformer_2023}
Zhang Youmin, Guo Xianda, Poggi Matteo, Zhu Zheng, Huang Guan, and Mattoccia
  Stefano.
\newblock {CompletionFormer}: {Depth} {Completion} with {Convolutions} and
  {Vision} {Transformers}, 2023.

\bibitem[Yu et~al.(2019)Yu, Lin, Yang, Shen, Lu, and Huang]{gate}
Jiahui Yu, Zhe Lin, Jimei Yang, Xiaohui Shen, Xin Lu, and Thomas~S Huang.
\newblock Free-form image inpainting with gated convolution.
\newblock In \emph{Proceedings of the IEEE/CVF international conference on
  computer vision}, pages 4471--4480, 2019.

\bibitem[Yuen et~al.(2011)Yuen, Yaoyuneyong, and Johnson]{yuen2011augmented}
Steve Chi-Yin Yuen, Gallayanee Yaoyuneyong, and Erik Johnson.
\newblock Augmented reality: An overview and five directions for ar in
  education.
\newblock \emph{Journal of Educational Technology Development and Exchange
  (JETDE)}, 4\penalty0 (1):\penalty0 11, 2011.

\bibitem[Yurtsever et~al.(2020)Yurtsever, Lambert, Carballo, and
  Takeda]{yurtsever2020survey}
Ekim Yurtsever, Jacob Lambert, Alexander Carballo, and Kazuya Takeda.
\newblock A survey of autonomous driving: Common practices and emerging
  technologies.
\newblock \emph{IEEE access}, 8:\penalty0 58443--58469, 2020.

\bibitem[Zhang(2012)]{zhang2012microsoft}
Zhengyou Zhang.
\newblock Microsoft kinect sensor and its effect.
\newblock \emph{IEEE multimedia}, 19\penalty0 (2):\penalty0 4--10, 2012.

\bibitem[Zhao et~al.(2021)Zhao, Gong, Fu, and Tao]{zhao_adaptive_2021}
Shanshan Zhao, Mingming Gong, Huan Fu, and Dacheng Tao.
\newblock Adaptive {Context}-{Aware} {Multi}-{Modal} {Network} for {Depth}
  {Completion}.
\newblock \emph{IEEE Transactions on Image Processing}, 30:\penalty0
  5264--5276, 2021.

\bibitem[Zhou et~al.(2023)Zhou, Yan, Liao, Lin, Huang, Zhao, Cui, and
  Li]{zhou_bev_2023}
Wending Zhou, Xu Yan, Yinghong Liao, Yuankai Lin, Jin Huang, Gangming Zhao,
  Shuguang Cui, and Zhen Li.
\newblock {BEV}@ {DC}: {Bird}'s-{Eye} {View} {Assisted} {Training} for {Depth}
  {Completion}.
\newblock In \emph{Proceedings of the {IEEE}/{CVF} {Conference} on {Computer}
  {Vision} and {Pattern} {Recognition}}, pages 9233--9242, 2023.

\end{thebibliography}
}

\clearpage
\setcounter{page}{1}
\maketitlesupplementary

\begin{strip}
    \centering
    \vspace{-2em}
    \centering
    \includegraphics[width=\linewidth]{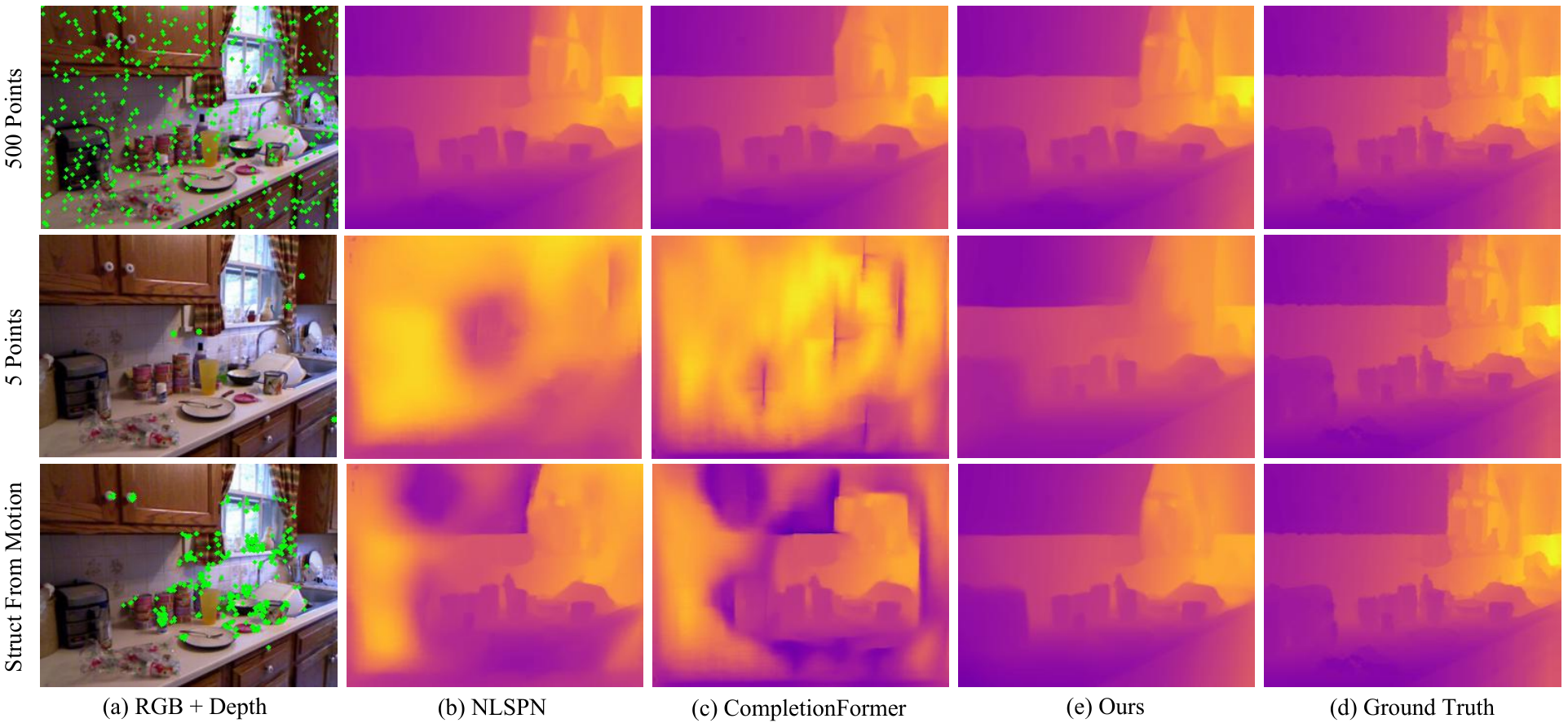}
    \vspace{-1em}
    \captionof{figure}{\textbf{Qualitative results on NYU Depth V2.} Comparing completion results with different patterns of inputs using different methods trained in fixed distribution.}
    \label{fig:a_nyu}
\end{strip}

\section{Appendix}
\label{sec:appendix}
In the following, we show additional experimental results.
\subsection{More results on NYU Depth V2}
We trained the comparison methods using their released code and the same training strategy with us (described in \cref{sec:dataset}), because the existing methods trained on fixed spatial patterns (500 random samples) are not comparable in the face of some challenging cases. To prove this point, we report the results of their pre-trained models in \cref{tab:nyu_pretrain}.
\begin{table}[!h]
    \centering
    \footnotesize
    \begin{tabular}{c|c|cc}
    \toprule
    Model            & Samples      & RMSE(m)$\downarrow$                                                                     & REL$\downarrow$                                    \\ \midrule
    NLSPN\cite{vedaldi_non-local_2020}            & 500          & 0.0917                                  & {\color[HTML]{FF0000} \textbf{0.0116}} \\
    CompletionFormer\cite{youmin_completionformer_2023} & 500          & {\color[HTML]{FF0000} \textbf{0.0901}}                                  & 0.0119                                 \\
    \textbf{Ours}    & 500          & 0.0976                                                                  & 0.0128                                 \\ \midrule
    NLSPN\cite{vedaldi_non-local_2020}            & 5            & 1.0389                                                                  & 0.2647                                 \\
    CompletionFormer\cite{youmin_completionformer_2023} & 5            & 1.1405                                                                 & 0.3073                                 \\
    \textbf{Ours}    & 5            & {\color[HTML]{FF0000} \textbf{0.3816}}  & {\color[HTML]{FF0000} \textbf{0.0877}} \\ \midrule
    NLSPN\cite{vedaldi_non-local_2020}            & ORB keypoint & 0.5997                                                          & 0.1462                                 \\
    CompletionFormer\cite{youmin_completionformer_2023} & ORB keypoint                                 & 0.6222                                 & 0.2222                                 \\
    \textbf{Ours}    & ORB keypoint & {\color[HTML]{FF0000} \textbf{0.2441}}  & {\color[HTML]{FF0000} \textbf{0.0541}} \\ \bottomrule
    \end{tabular}
    \caption{\textbf{Results on NYU Depth V2}. Quantitative comparisons with the state-of-the-art methods trained in fixed distribution. The best results are red and in bold.}
    \label{tab:nyu_pretrain}
\end{table} 
The visualization results in \cref{fig:a_nyu} also illustrate that the existing methods cannot reconstruct the overall scene structure when dealing with such challenging cases.

Furthermore, we validated the performance of our method with a fixed distribution. We used 500 sampling random depth points as inputs and retrained our model, like the previous setting. Even though SparseDC is focused on depth completion from \textit{sparse and non-uniform} inputs, our method still exhibits comparable performance as shown in the \cref{tab:nyu2}.
\begin{table}[t]
    \centering
    \footnotesize
    \begin{tabular}{ccccccc}
    \toprule
    Methods          & Params & RMSE(m)$\downarrow$ & REL$\downarrow$                                   \\ \midrule
GuideNet\cite{tang_learning_2019}         & 74M    & 0.101   & 0.015                                 \\
DySPN\cite{lin_dynamic_2022}            & 26M    & {\color[HTML]{FF0000}\textbf{0.090}}   & {\color[HTML]{FF0000} \textbf{0.012}} \\
NLSPN\cite{vedaldi_non-local_2020}            & 25.8M   & 0.092   & {\color[HTML]{FF0000} \textbf{0.012}} \\
GraphCSPN\cite{liu_graphcspn_2022}        & 26M    & {\color[HTML]{FF0000}\textbf{0.090}}    & {\color[HTML]{FF0000} \textbf{0.012}} \\
CompletionFormer\cite{youmin_completionformer_2023} & 45M    & {\color[HTML]{FF0000}\textbf{0.090}}    & {\color[HTML]{FF0000} \textbf{0.012}} \\ \midrule
Ours             & 38.2M  & 0.093   & {\color[HTML]{FF0000} \textbf{0.012}}  \\  \bottomrule
    \end{tabular}
    \caption{\textbf{Results on NYU Depth V2 with fixed distribution.} Quantitative comparisons with state-of-the-art methods in a fixed spatial pattern.}
    \label{tab:nyu2}
\end{table}

\begin{figure*}[p]
    \centering
    \includegraphics[width=1.0\linewidth]{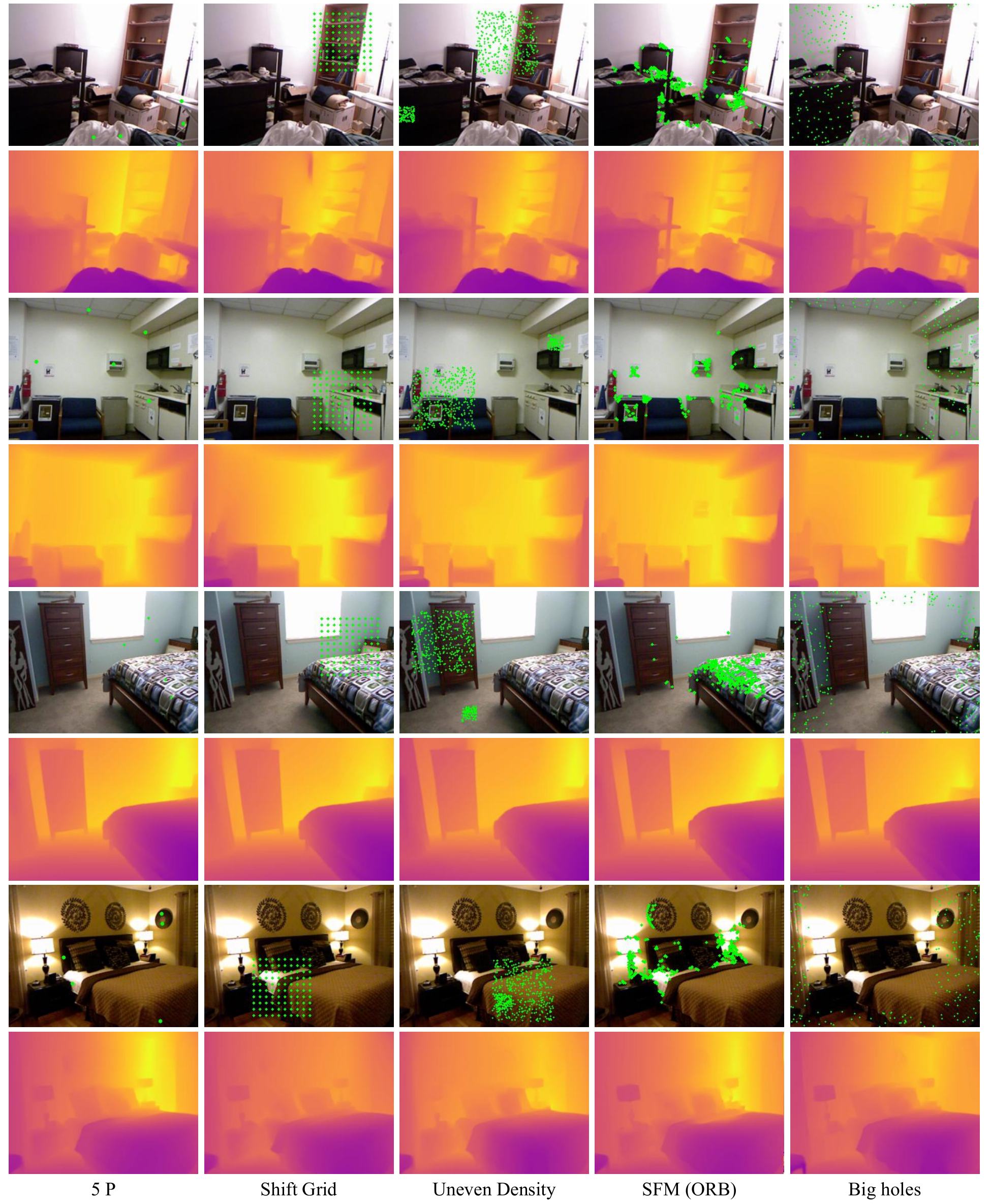}
    \caption{\textbf{Qualitative results on NYU Depth V2.} Examples of using SparseDC to complete sparse depth with different patterns.}
    \label{fig:a_nyu2}
\end{figure*}

Finally, we report the performance of comparison methods using other metrics in \cref{tab:nyu}, and show more visualizations in \cref{fig:a_nyu2}.
\begin{table*}[t]
    \centering
    \footnotesize
    \begin{tabular}{@{}|c|c|cccccccc|@{}}
        \toprule
        Methods                                               & Samples      & RMSE (m)$\downarrow$                   & MAE (m)$\downarrow$                    & IRMSE ($\frac{1}{m}$)$\downarrow$                  & iMAE ($\frac{1}{m}$)$\downarrow$                   & REL $\downarrow$                    & $\delta_{1.25}$  $\uparrow$                         & $\delta_{1.25}^2$ $\uparrow$                          & $\delta_{1.25}^3$ $\uparrow$                          \\ \midrule
        NLSPN\cite{vedaldi_non-local_2020}                  & 500          & 0.0984                                 & 0.0395                                 & 0.0153                                 & 0.0058                                 & 0.0136                                 & {\color[HTML]{FF0000} \textbf{0.9949}} & {\color[HTML]{FF0000} \textbf{0.9992}} & {\color[HTML]{FF0000} \textbf{0.9998}} \\
        GraphCSPN\cite{liu_graphcspn_2022}                  & 500          & 0.0990                                 & 0.0399                                 & 0.0153                                 & 0.0059                                 & 0.0136                                 & 0.9948                                 & {\color[HTML]{FF0000} \textbf{0.9992}} & {\color[HTML]{FF0000} \textbf{0.9998}} \\
        CompletionFormer\cite{youmin_completionformer_2023} & 500          & 0.0984                                 & 0.0392                                 & 0.0153                                 & 0.0058                                 & 0.0135                                 & 0.9948                                 & {\color[HTML]{FF0000} \textbf{0.9992}} & {\color[HTML]{FF0000} \textbf{0.9998}} \\
        \textbf{Ours}                                                & 500          & {\color[HTML]{FF0000} \textbf{0.0976}} & {\color[HTML]{FF0000} \textbf{0.0379}} & {\color[HTML]{FF0000} \textbf{0.0148}} & {\color[HTML]{FF0000} \textbf{0.0054}} & {\color[HTML]{FF0000} \textbf{0.0128}} & {\color[HTML]{FF0000} \textbf{0.9949}} & {\color[HTML]{FF0000} \textbf{0.9992}} & {\color[HTML]{FF0000} \textbf{0.9998}} \\ \midrule
        NLSPN\cite{vedaldi_non-local_2020}                  & 200          & {\color[HTML]{FF0000} \textbf{0.1336}} & 0.0579                                 & 0.0209                                 & 0.0087                                 & 0.0201                                 & {\color[HTML]{FF0000} \textbf{0.9902}} & {\color[HTML]{FF0000} \textbf{0.9983}} & {\color[HTML]{FF0000} \textbf{0.9996}} \\
        GraphCSPN\cite{liu_graphcspn_2022}                  & 200          & 0.1338                                 & 0.0575                                 & 0.0207                                 & 0.0085                                 & 0.0197                                 & {\color[HTML]{FF0000} \textbf{0.9902}} & 0.9982                                 & 0.9995                                 \\
        CompletionFormer\cite{youmin_completionformer_2023} & 200          & 0.1349                                 & 0.0588                                 & 0.0214                                 & 0.0089                                 & 0.0205                                 & 0.9898                                 & 0.9982                                 & 0.9995                                 \\
        \textbf{Ours}                                                & 200          & 0.1346                                 & {\color[HTML]{FF0000} \textbf{0,0565}} & {\color[HTML]{FF0000} \textbf{0.0203}} & {\color[HTML]{FF0000} \textbf{0.0081}} & {\color[HTML]{FF0000} \textbf{0.0191}} & 0.9899                                 & 0.9982                                 & 0,9995                                 \\ \midrule
        NLSPN\cite{vedaldi_non-local_2020}                  & 50           & 0.2123                                 & 0.1078                                 & 0.0345                                 & 0.0168                                 & 0.0382                                 & 0.9743                                 & {\color[HTML]{FF0000} \textbf{0.9948}} & {\color[HTML]{FF0000} \textbf{0.9986}} \\
        GraphCSPN\cite{liu_graphcspn_2022}                  & 50           & 0.2119                                 & 0.1069                                 & 0.0340                                 & 0.0165                                 & 0.0375                                 & 0.9745                                 & 0.9942                                 & 0.9984                                 \\
        CompletionFormer\cite{youmin_completionformer_2023} & 50           & 0.2183                                 & 0.1124                                 & 0.0362                                 & 0.0178                                 & 0.0405                                 & 0.9717                                 & 0.9941                                 & 0.9984                                 \\
        \textbf{Ours}                                                & 50           & {\color[HTML]{FF0000} \textbf{0.2079}} & {\color[HTML]{FF0000} \textbf{0.1017}} & {\color[HTML]{FF0000} \textbf{0.0323}} & {\color[HTML]{FF0000} \textbf{0.0152}} & {\color[HTML]{FF0000} \textbf{0.0350}} & {\color[HTML]{FF0000} \textbf{0.9755}} & 0.9946                                 & 0.9985                                 \\ \midrule
        NLSPN\cite{vedaldi_non-local_2020}                  & 10           & 0.3639                                 & 0.2254                                 & 0.0624                                 & 0.0373                                 & 0.0827                                 & 0.9154                                 & 0.9810                                 & 0.9950                                 \\
        GraphCSPN\cite{liu_graphcspn_2022}                  & 10           & 0.3650                                 & 0.2237                                 & 0.0639                                 & 0.0373                                 & 0.0832                                 & 0.9138                                 & 0.9796                                 & 0.9943                                 \\
        CompletionFormer\cite{youmin_completionformer_2023} & 10           & 0.3642                                 & 0.2300                                 & 0.0650                                 & 0.0387                                 & 0.0874                                 & 0.9110                                 & 0.9802                                 & 0.9946                                 \\
        \textbf{Ours}                                                & 10           & {\color[HTML]{FF0000} \textbf{0.3226}} & {\color[HTML]{FF0000} \textbf{0.1923}} & {\color[HTML]{FF0000} \textbf{0.0536}} & {\color[HTML]{FF0000} \textbf{0.0308}} & {\color[HTML]{FF0000} \textbf{0.0681}} & {\color[HTML]{FF0000} \textbf{0.9358}} & {\color[HTML]{FF0000} \textbf{0.9854}} & {\color[HTML]{FF0000} \textbf{0.9960}} \\ \midrule
        NLSPN\cite{vedaldi_non-local_2020}                  & 5            & 0.4439                                 & 0.2953                                 & 0.0770                                 & 0.0497                                 & 0.1094                                 & 0.8674                                 & 0.9696                                 & 0.9925                                 \\
        GraphCSPN\cite{liu_graphcspn_2022}                  & 5            & 0.4555                                 & 0.3001                                 & 0.0814                                 & 0.0512                                 & 0.1151                                 & 0.8604                                 & 0.9650                                 & 0.9899                                 \\
        CompletionFormer\cite{youmin_completionformer_2023} & 5            & 0.4377                                 & 0.2961                                 & 0.0791                                 & 0.0507                                 & 0.1144                                 & 0.8657                                 & 0.9683                                 & 0.9917                                 \\
        \textbf{Ours}                                                & 5            & {\color[HTML]{FF0000} \textbf{0.3816}} & {\color[HTML]{FF0000} \textbf{0.2456}} & {\color[HTML]{FF0000} \textbf{0.0649}} & {\color[HTML]{FF0000} \textbf{0.0405}} & {\color[HTML]{FF0000} \textbf{0.0877}} & {\color[HTML]{FF0000} \textbf{0.9059}} & {\color[HTML]{FF0000} \textbf{0.9795}} & {\color[HTML]{FF0000} \textbf{0.9945}} \\ \midrule
        NLSPN\cite{vedaldi_non-local_2020}                  & shift\_grid  & 0.4263                                 & 0.2718                                 & 0.0791                                 & 0.0487                                 & 0.1043                                 & 0.8655                                 & 0.9671                                 & 0.9923                                 \\
        GraphCSPN\cite{liu_graphcspn_2022}                  & shift\_grid  & 0.4485                                 & 0.2808                                 & 0.0833                                 & 0.0502                                 & 0.1083                                 & 0.8531                                 & 0.9614                                 & 0.9899                                 \\
        CompletionFormer\cite{youmin_completionformer_2023} & shift\_grid  & 0.4313                                 & 0.2812                                 & 0.0834                                 & 0.0512                                 & 0.1149                                 & 0.8560                                 & 0.9635                                 & 0.9901                                 \\
        \textbf{Ours}                                                & shift\_grid  & {\color[HTML]{FF0000} \textbf{0.3736}} & {\color[HTML]{FF0000} \textbf{0.2331}} & {\color[HTML]{FF0000} \textbf{0.0680}} & {\color[HTML]{FF0000} \textbf{0.0409}} & {\color[HTML]{FF0000} \textbf{0.0852}} & {\color[HTML]{FF0000} \textbf{0.8987}} & {\color[HTML]{FF0000} \textbf{0.9782}} & {\color[HTML]{FF0000} \textbf{0.9947}} \\ \midrule
        NLSPN\cite{vedaldi_non-local_2020}                  & uneven density & 0.3931                                 & 0.2438                                 & 0.0710                                 & 0.0425                                 & 0.0909                                 & 0.8901                                 & 0.9741                                 & 0.9935                                 \\
        GraphCSPN\cite{liu_graphcspn_2022}                  & uneven density & 0.3991                                 & 0.2420                                 & 0.0737                                 & 0.0427                                 & 0.0920                                 & 0.8855                                 & 0.9705                                 & 0.9922                                 \\
        CompletionFormer\cite{youmin_completionformer_2023} & uneven density & 0.3919                                 & 0.2432                                 & 0.0724                                 & 0.0424                                 & 0.0961                                 & 0.8867                                 & 0.9719                                 & 0.9921                                 \\
        \textbf{Ours}                                                & uneven density & {\color[HTML]{FF0000} \textbf{0.3543}} & {\color[HTML]{FF0000} \textbf{0.2141}} & {\color[HTML]{FF0000} \textbf{0.0632}} & {\color[HTML]{FF0000} \textbf{0.0372}} & {\color[HTML]{FF0000} \textbf{0.0763}} & {\color[HTML]{FF0000} \textbf{0.9121}} & {\color[HTML]{FF0000} \textbf{0.9809}} & {\color[HTML]{FF0000} \textbf{0.9950}} \\ \midrule
        NLSPN\cite{vedaldi_non-local_2020}                  & ORB keypoint & 0.2590                                 & 0.1565                                 & 0.0542                                 & 0.0301                                 & 0.0625                                 & 0.9479                                 & 0.9890                                 & 0.9965                                 \\
        GraphCSPN\cite{liu_graphcspn_2022}                  & ORB keypoint & 0.2571                                 & 0.1510                                 & 0.0540                                 & 0.0290                                 & 0.0603                                 & 0.9481                                 & 0.9880                                 & 0.9961                                 \\
        CompletionFormer\cite{youmin_completionformer_2023} & ORB keypoint & 0.2660                                 & 0.1616                                 & 0.0576                                 & 0.0316                                 & 0.0672                                 & 0.9417                                 & 0.9873                                 & 0.9965                                 \\
        \textbf{Ours}                                                & ORB keypoint & {\color[HTML]{FF0000} \textbf{0.2441}} & {\color[HTML]{FF0000} \textbf{0.1404}} & {\color[HTML]{FF0000} \textbf{0.0498}} & {\color[HTML]{FF0000} \textbf{0.0265}} & {\color[HTML]{FF0000} \textbf{0.0541}} & {\color[HTML]{FF0000} \textbf{0.9539}} & {\color[HTML]{FF0000} \textbf{0.9898}} & {\color[HTML]{FF0000} \textbf{0.9970}} \\ \midrule
         NLSPN\cite{vedaldi_non-local_2020}                  & big holes          & 0.2368 & 0.1205                                 & 0.0332                                 & 0.0171                                 & 0.0402                                 & 0.9695                                 & 0.9944 & 0.9986 \\
        GraphCSPN\cite{liu_graphcspn_2022}                  & big holes          & 0.2443                                 & 0.1224                                 & 0.0336                                 & 0.0171                                 & 0.0403                                 & 0.9675 & 0.9932                                 & 0.9983                                 \\
        CompletionFormer\cite{youmin_completionformer_2023} & big holes          & 0.2493                                 & 0.1273                                 & 0.0350                                 & 0.0180                                 & 0.0434                                 & 0.9648                                 & 0.9933                                 & 0.9983                                 \\
        \textbf{Ours}                                                & big holes          & {\color[HTML]{FF0000} \textbf{0.2262}}                                 & {\color[HTML]{FF0000} \textbf{0.1121}} & {\color[HTML]{FF0000} \textbf{0.0307}} & {\color[HTML]{FF0000} \textbf{0.0155}} & {\color[HTML]{FF0000} \textbf{0.0362}} & {\color[HTML]{FF0000} \textbf{0.9731}}                                 & {\color[HTML]{FF0000} \textbf{0.9944}}                                 & {\color[HTML]{FF0000} \textbf{0.9985}}                                 \\
        \bottomrule
    \end{tabular}
    \caption{\textbf{Results on NYU Depth V2 in all Metrics.} Quantitative comparisons with state-of-the-art methods in different spatial patterns. The best results are red and in bold.}
    \label{tab:nyu}
\end{table*}

\subsection{More results on KITTI}
\begin{figure*}[p]
    \centering
    \includegraphics[width=1.0\linewidth]{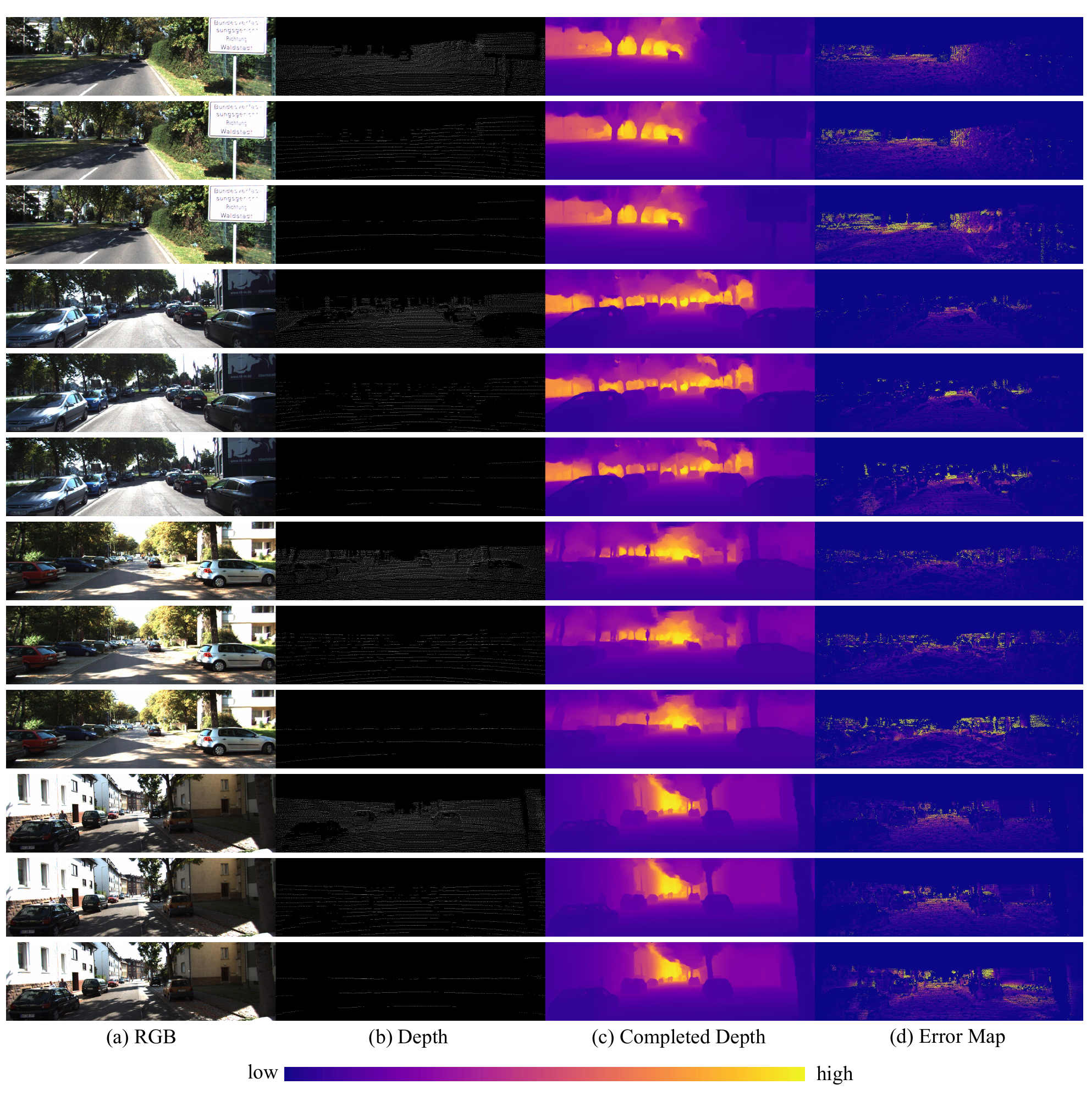}
    \caption{\textbf{Qualitative results on Kitti.} We report completed results from SparseDC with 4, 16, and 64 line depth inputs, respectively.}
    \label{fig:a_kitti}
\end{figure*}

\cref{fig:a_kitti} shows, on an image of the KITTI DC dataset and three different depth inputs, the outcome of SparseDC. Our method can effectively maintain the overall scene structure under different lines of depth inputs.

\subsection{More results on SUN RGB-D}
\begin{figure*}[p]
    \centering
    \includegraphics[width=1.0\linewidth]{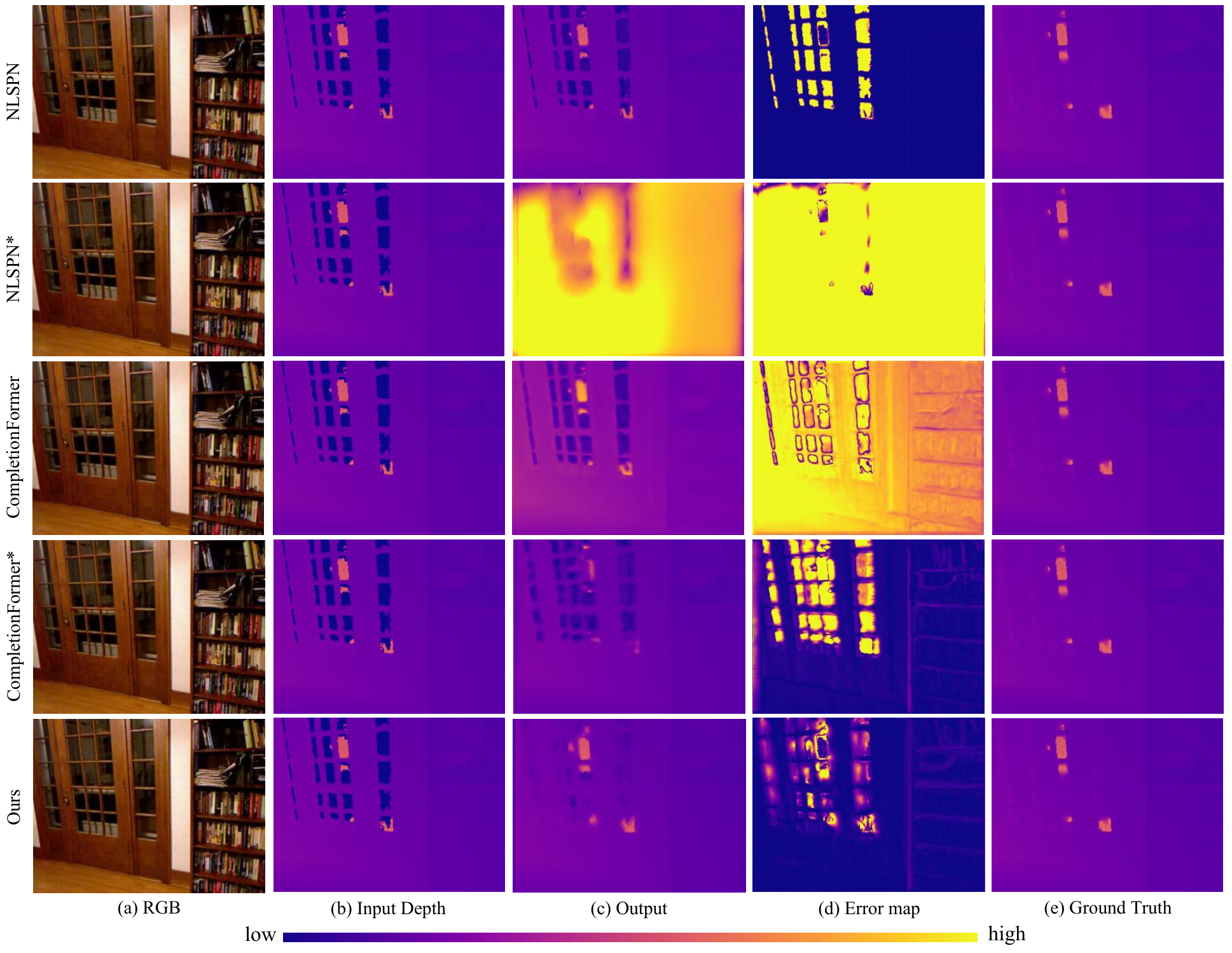}
    \caption{\textbf{Qualitative results on SUNRGBD.} $*$ denotes trained model using our training strategy, without $*$ is their released pretrained model trained using fixed patterns.}
    \label{fig:sunrgbd}
\end{figure*}

SUN RGB-D\cite{song2015sun} contains 10,355 RGB-D images captured by four different sensors. It was used to validate the generalization ability of our method. We used the noisy and incomplete raw depth data as inputs without sampling, and used the reﬁned depth maps based on multiple frames as the ground truths for evaluation. The input images were resized to 320$\times$240 and randomly cropped to 304$\times$228. The Qualitative comparison results are shown in \cref{fig:sunrgbd}. For such challenging cases (noisy, incomplete, unseen), existing methods with or without augment-based strategy are difficult to obtain satisfactory results, whereas our method achieved the best performance due to its meticulous design. 

\end{document}